\documentclass[lettersize,journal]{IEEEtran}
\usepackage{amsmath,amsfonts}
\usepackage{algorithmic}
\usepackage{algorithm}
\usepackage{array}
\usepackage[caption=false,font=footnotesize,labelfont=rm,textfont=rm]{subfig}
\usepackage{textcomp}
\usepackage{stfloats}
\usepackage{url}
\usepackage{verbatim}
\usepackage{graphicx}
\usepackage{cite}
\usepackage{multirow}
\usepackage{booktabs}
\usepackage{courier}
\usepackage{color}
\usepackage{bbm}
\usepackage[table,xcdraw]{xcolor}
\usepackage[backref]{hyperref}
\hypersetup{hypertex=true,
	colorlinks=true,
	linkcolor=blue,
	anchorcolor=blue,
	citecolor=blue
}
\usepackage[font=footnotesize,labelfont=rm,textfont=rm]{caption}
\begin{document}

%\title{Efficient Multispectral Object Detector: Rethinking Early-Fusion Strategies for Improved Performance}
\title{Rethinking Early-Fusion Strategies for Improved Multispectral Object Detection}

\author{Xue Zhang, Si-Yuan Cao, Fang Wang, Runmin Zhang, Zhe Wu, Xiaohan Zhang, Xiaokai Bai, and \\Hui-Liang Shen, \emph{Senior Member, IEEE}
\thanks{
This work was supported in part by the National Key Research and Development Program of China under grant 2023YFB3209800, in part by the Natural Science Foundation of Zhejiang Province under grant D24F020006, in part by the National Natural Science Foundation of China under grant 62301484, and in part by the Jinhua Science and Technology Bureau Project. \emph{(Corresponding authors: Si-Yuan Cao and Hui-Liang Shen.)}	
	
X. Zhang, R. Zhang, Z. Wu, X. Zhang, and X. Bai are with the College of Information Science and Electronic Engineering, Zhejiang University, Hangzhou 310027, China (e-mail: zxue2019@zju.edu.cn, runmin\_zhang@zju.edu.cn, jeffw@zju.edu.cn, zhangxh2023@zju.edu.cn, shawnnnkb@zju.edu.cn).

S.-Y. Cao is with the Ningbo Research Institute, College of Information Science and Electronic Engineering, Zhejiang University, China (e-mail: cao\_siyuan@zju.edu.cn).

F. Wang is with the School of Information and Electrical Engineering, Hangzhou City University, and also with the Hangzhou City University Binjiang Innovation Center, China (e-mail: wangf@zucc.edu.cn)

H.-L. Shen is with the College of Information Science and Electronic Engineering, Zhejiang University, the Jinhua Institute of Zhejiang University, and the Key Laboratory of Collaborative Sensing and Autonomous Unmanned Systems of Zhejiang Province, China (e-mail: shenhl@zju.edu.cn).

}}

%% The paper headers
%% Remember, if you use this you must call \IEEEpubidadjcol in the second
%% column for its text to clear the IEEEpubid mark.

\maketitle

\begin{abstract}
Most recent multispectral object detectors employ a two-branch structure to extract features from RGB and thermal images. While the two-branch structure achieves better performance than a single-branch structure, it overlooks inference efficiency. This conflict is increasingly aggressive, as recent works solely pursue higher performance rather than both performance and efficiency. In this paper, we address this issue by improving the performance of efficient single-branch structures. We revisit the reasons causing the performance gap between these structures. For the first time, we reveal the information interference problem in the naive early-fusion strategy adopted by previous single-branch structures. Besides, we find that the domain gap between multispectral images, and weak feature representation of the single-branch structure are also key obstacles for performance. Focusing on these three problems, we propose corresponding solutions, including a novel shape-priority early-fusion strategy, a weakly supervised learning method, and a core knowledge distillation technique. Experiments demonstrate that single-branch networks equipped with these three contributions achieve significant performance enhancements while retaining high efficiency. Our code will be available at \url{https://github.com/XueZ-phd/Efficient-RGB-T-Early-Fusion-Detection}.

\end{abstract}


 \begin{figure*}[t]
	\centering
	\subfloat{\includegraphics[width=0.65\linewidth]{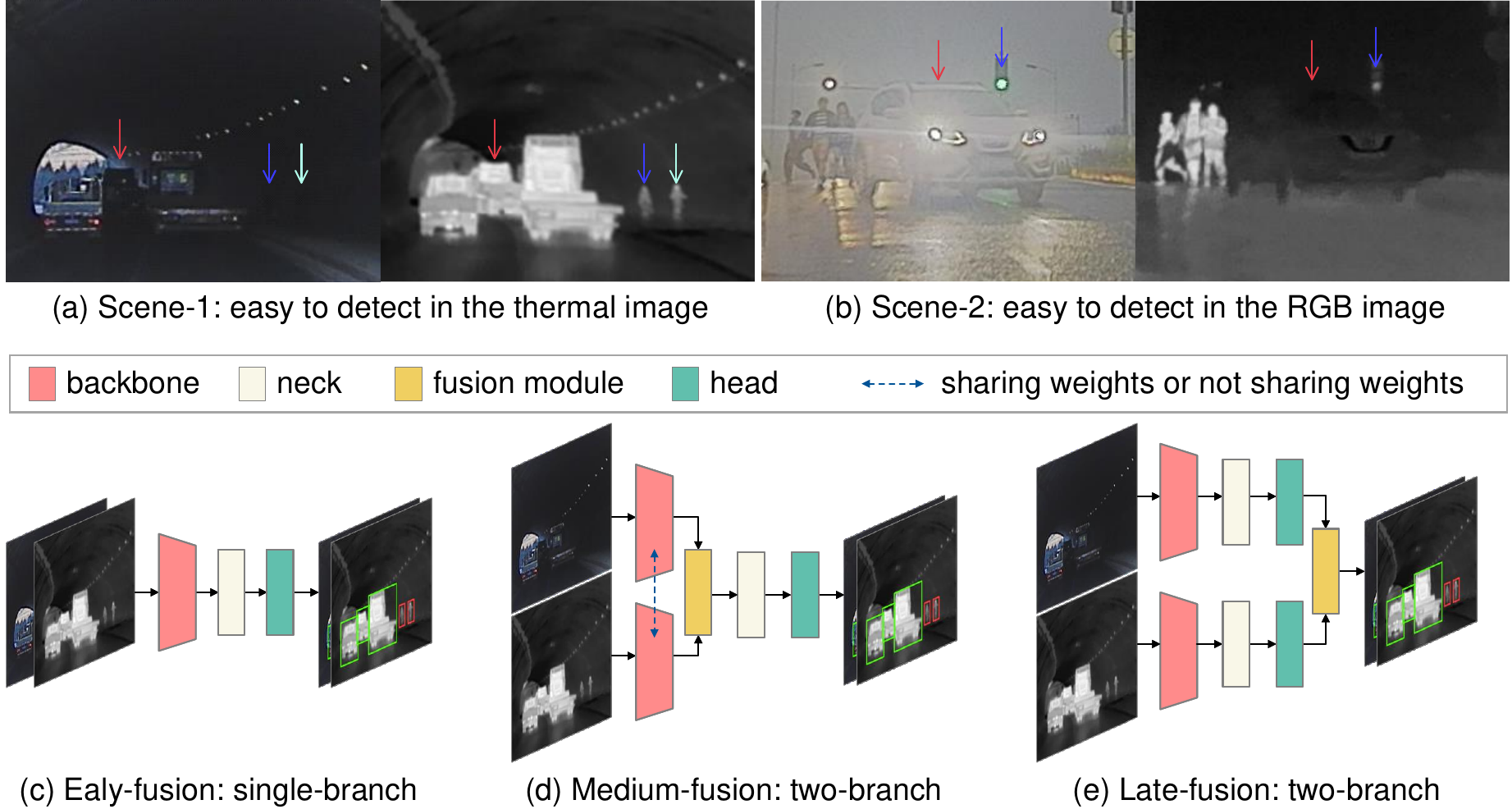}}
	\subfloat{\includegraphics[width=0.35\linewidth]{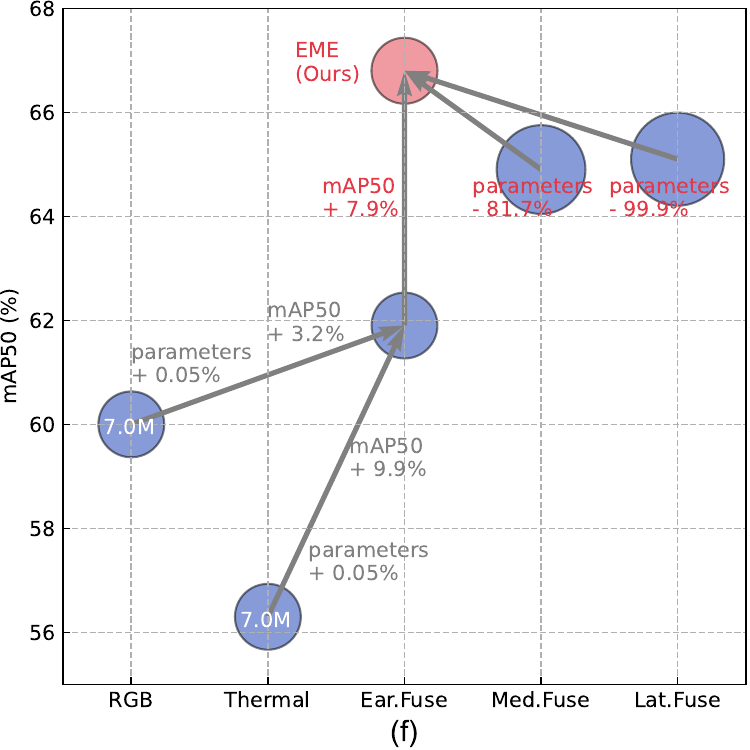}}
	\caption{Multispectral object detection and fusion strategies. (a) In Scene-1, objects are easier to detect in the thermal image. (b) In Scene-2, objects are easier to detect in the RGB image. (c) Early-fusion strategy. (d) Medium-fusion strategy. (e) Late-fusion strategy. (f) Detection results of different strategies on the M3FD dataset~\cite{liu2022target}. YOLOv5~\cite{yolov5} is adopted as the baseline in this experiment. The area of each circle denotes the number of parameters.}
	\label{fig:background}
 	\vspace{-5mm}
\end{figure*}

\section{Introduction}
% Background: multispectral object detection
\IEEEPARstart{M}{ultispectral} object detection has been widely studied, since multispectral images can provide complementary information to achieve consistent detection in various lighting conditions~\cite{pdmsc, CACFNet, RBIT, farooq2022evaluation, TISPT, MTANet, umssde, qafa}. This complementarity is illustrated in Fig.~\ref{fig:background}~(a) and (b). Given the multispectral inputs, modern multispectral detectors develop three fusion strategies: early-fusion, medium-fusion, and late-fusion, shown in Fig.~\ref{fig:background}~(c)~-~(e). The medium and late-fusion strategies often achieve superior performance compared to early-fusion~\cite{liu2016multispectral, wagner2016multispectral, xie2023illumination, RBIT, CACFNet, liu2021deep}. However, they use a two-branch structure, making model deployment on edge devices expensive.
% Conflict1: two-branch vs. single-branch
In contrast, the early-fusion strategy adopts a simple single-branch structure, facilitating deployment on edge devices. Nevertheless, its performance is low, and there are few works to address this problem, resulting in an increasing gap between high performance and high efficiency.

% The root cause
{The main motivation for this work is to resolve the conflict between detection performance and inference efficiency. To this end, we focus on improving the performance of the early-fusion strategy while maintaining its high computational efficiency.} We first conduct pilot studies and observe that a plain early-fusion strategy cannot consistently obtain improved performances compared to single-modality inputs. Based on this observation, we rethink the early-fusion strategy and summarize three key obstacles: 1) the information interference problem when simply concatenating the multispectral images, 2) the domain gap existing in thermal and RGB images, and 3) the weak feature representation of the single-branch structure. Focusing on these obstacles, we propose corresponding solutions.\\
\textbf{- Information interference problem} refers to the potential suppression of important information in one modality by another. In the plain early-fusion strategy, previous works~\cite{zhang2021guided} typically feed concatenated multispectral images into a convolution layer and generate a fused feature. The convolution layer generally has a small receptive field. Therefore, based on limited contexts, this approach is hard to determine which modality information is important. We address this issue by first recognizing that object shapes are agnostic to visible and infrared wavelengths and devise a module to fuse multispectral images based on object shape saliency, named the \underline{sha}pe-\underline{p}riority \underline{e}arly-fusion (ShaPE) module.\\
\textbf{- Domain gap between RGB and thermal images} is usually neglected in previous works. They generally adopt an RGB pre-trained backbone network to extract features from both RGB and thermal images~\cite{xie2023illumination, RBIT}. However, the domain gap may cause the representation distribution shift. This issue is also recognized in the work~\cite{dformer} on an RGB-D task. Different from previous works, we introduce a weakly supervised learning method to address this issue. Within this method, the backbone network jointly uses RGB and thermal images to learn the representation of CLIP~\cite{clip}, since CLIP has demonstrated promising zero-shot generalizability in bridging the domain gap~\cite{PADCLIP}. Additionally, we introduce a segmentation auxiliary branch. Our method allows the backbone network to reduce representation shifts and improve semantic localization ability.\\
\textbf{- Weak feature representation problem} results from the early-fusion strategy employing a single-branch structure. This structure has fewer parameters and simpler fusion modules compared to medium and late-fusion strategies. We address this issue by introducing the knowledge distillation (KD) technique~\cite{hinton2015distilling}. In KD, a key problem is how to align the feature dimensions between teacher and student models. Previous works generally introduce a convolution layer for the student model to learn all knowledge from the teacher model~\cite{fitnet, chen2017learning}. However, we show that not all information in teacher model is helpful for downstream tasks. Therefore, we introduce core knowledge distillation (CoreKD) to transfer the most crucial knowledge for specific downstream tasks, resembling the human learning process where the teacher highlights key knowledge for quick understanding and absorption by the students.

Experimental results validate that our efficient multispectral early-fusion (EME) detector achieves a significant performance improvement without considerably increasing the number of parameters, as shown in Fig.~\ref{fig:background}~(f). Besides, our EME outperforms the previous state-of-the-art approaches. In summary, our contributions are threefold:

\begin{itemize}
\item {We systematically analyze the causes of the performance gap between single-branch and two-branch structures. Unlike previous works, we identify and summarize three key obstacles limiting the single-branch early-fusion strategy: information interference, domain gap, and weak feature representation. Notably, information interference between multispectral images is revealed for the first time in this work.}

\item For each obstacle, we propose the corresponding solution: we develop 1) a ShaPE module to address the information interference issue, 2) a weakly supervised learning method to reduce domain gap and improve semantic localization abilities, and 3) a CoreKD to enhance the feature representation of single-branch networks.

\item Extensive experiments validate that the early-fusion strategy, equipped with our ShaPE module, weakly supervised learning, and CoreKD technique, demonstrates significant improvement. {These three modules benefit various common detectors, such as YOLOv5~\cite{yolov5}, RetinaNet~\cite{lin2017focal}, and GFL~\cite{gfl}. Importantly, only the ShaPE module is retained during the inference phase. Consequently, our method achieves both high performance and efficiency.}
\end{itemize}

\section{Related Work}
 In this section, we offer a brief overview of multispectral object detection and introduce related works in weakly supervised learning and knowledge distillation.

\subsection{Multispectral Object Detection}\label{sec:relatedmultispectralod}
Multi-source information fusion~\cite{Flexible-Mixup, mifn, slanet, csma} has exhibited promising application potential in computer vision tasks. In this work, we focus on the multispectral object detection task that uses RGB and thermal image pairs to detect objects.
According to fusion strategies, multispectral object detection can be classified into three categories: early-fusion, medium-fusion and late-fusion strategies. Previous works~\cite{liu2016multispectral}, \cite{wagner2016multispectral} and \cite{proben} confirm that both medium-fusion and late-fusion strategies outperform the early-fusion strategy.

However, both the medium and late fusion strategies adopt a two-branch structure that limits their use on resource-limited edge devices. Previous works notice this weakness and provide some solutions. For example, in \cite{liu2021deep}, a model using the medium-fusion strategy is first trained as a teacher, and its knowledge is transferred to a student model. The student model only receives RGB images as inputs. Although it saves resources, it discards important complementary information from thermal images. The work~\cite{xie2023illumination} introduces a domain adaptation technique. It uses a medium-fusion model to guide single-branch model learning, which only receives thermal images as inputs and also discards complementary information from RGB images. To employ complementary information while saving computational resources, \cite{zhang2022low} transfers knowledge from a medium-fusion model to an early-fusion model. Nevertheless, it neglects information interference problem. Some works in the image fusion field~\cite{liu2022target, CDDFuse, DIVFusion} demonstrate that fused images can improve detectors, but the fusion process still introduces an additional computational burden.

Different from previous works, we identify the information interference problem in early-fusion strategies. By addressing this problem, we fully employ the complementary information in multispectral images, without significantly increasing computational burden.

\subsection{Weakly Supervised Learning in Object Detection}\label{sec:relatedweaklysupervised}
Weakly supervised learning has received much attention in object localization and
detection, as comprehensively surveyed in \cite{zhang2021weakly}. Recent works in the multispectral object detection adopt this technique. Based on the weak annotations they utilize, we can coarsely divide them into image- and box-level weakly supervised learning approaches.

In image-level weakly supervised learning approaches, previous works mainly employ the illumination condition of RGB images as weighting factors to determine the modality importance~\cite{liu2021deep, RBIT, zhang2023illumination, xie2023illumination}. In box-level approaches, previous works~\cite{zhang2021guided, zhang2023tfdet} mainly employ the bounding-box annotations to generate masks. They use these masks to construct spatial attention mechanisms, highlighting representations within target regions.

Different from previous works, we use weakly supervised learning to address the domain gap problem in RGB and thermal images. We employ image-level labels to construct a multi-label classification auxiliary task. This task can fully exploit the complementary information in multispectral images, instead of solely using information from one modality. Along with the powerful CLIP model~\cite{clip} and box-level weak labels, our method can reduce the domain gap and obtain precise semantic localization abilities.

\subsection{Knowledge Distillation}\label{sec:relatedkd}
Knowledge distillation is first introduced in~\cite{hinton2015distilling}. It aims to improve a lightweight student model by learning knowledge from a high-capacity teacher model. According to distillation approaches, this technique can be roughly divided into two groups: logit distillation~\cite{hinton2015distilling} and feature distillation~\cite{fitnet}. The former let a student model learn the logit of a teacher model, while the latter let a student model learn the feature of a teacher model. These distillation approaches are also applied to object detection~\cite{li2023object, chen2017learning}. Recently, some works in multispectral object detection also employ the knowledge distillation technique~\cite{zhang2022low, liu2021deep}. In the distillation process, they generally introduce a projection layer to align the teacher and student feature channel number. The purpose of this approach is to learn all representations in the teacher model.

Different from previous works, we first confirm that not all information in teacher features is beneficial to downstream task including classification and regression. Based on this, we propose a core knowledge distillation technique to transfer the most important features for the downstream tasks to the student model.

\section{Method}
\begin{figure}[t]
	\centering
	\includegraphics[width=\linewidth]{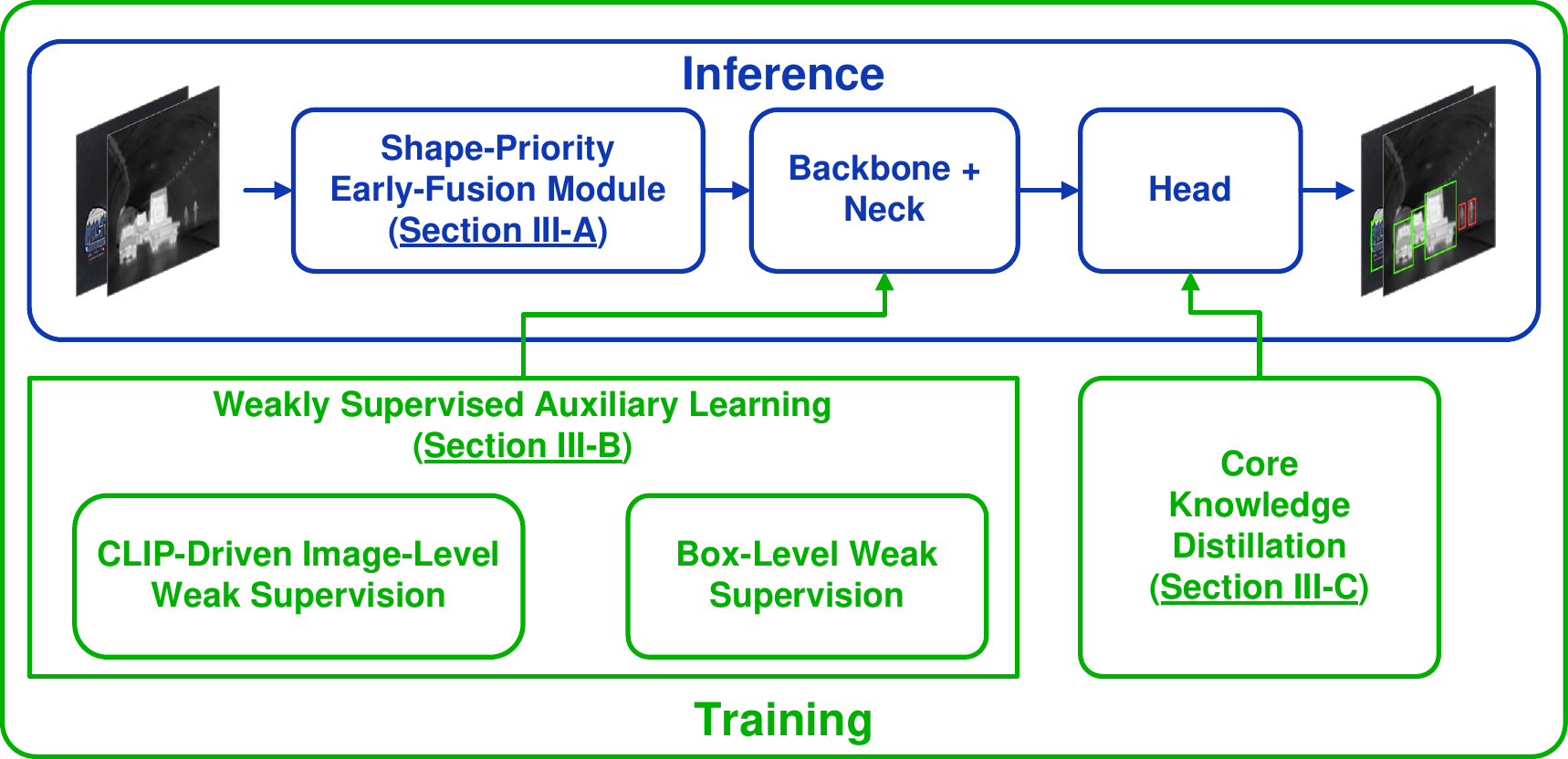}
	\caption{Overview of our method. We adopt the single-branch structure as the baseline model and develop three key modules: \underline{sha}pe-\underline{p}riority \underline{e}arly-fusion (ShaPE), weakly supervised auxiliary learning, and core knowledge distillation. The ShaPE module remains in both the inference and training phases, while the other two modules are removed in the inference phase.}
	\label{fig:method_overview}
	\vspace{-4mm}
\end{figure}

Fig.~\ref{fig:method_overview} illustrates the overview of our method, {where the training process and the inference process are marked in green and blue, respectively.} We adopt a single-branch structure as the baseline model considering its low memory cost. To boost its performance, we develop three key modules: \underline{sha}pe-\underline{p}riority \underline{e}arly-fusion (ShaPE), weakly supervised auxiliary learning, and core knowledge distillation (CoreKD). {Note that only the training process requires weakly supervised auxiliary learning and CoreKD, and both are removed during the inference phase. Consequently, our method adds only the ShaPE module to the early-fusion single-branch structure during the inference phase.} In the following sections, we describe the ShaPE module in Section~\ref{sec:ShaPE}, the weakly supervised auxiliary learning method in Section~\ref{sec:weakly}, and CoreKD in Section~\ref{sec:CoreKD}.

%To understand the effect of each contribution, we present necessary experimental results in this section. For descriptions of datasets and experimental details, please refer to Section~\ref{sec:setup}.

% Since we draw our insight from the observation on class-specific confidence score map of a detector, we first introduce its concept in Section~\ref{sec:preliminary}. Then, we describe our motivation in Section~\ref{sec:motivation}, and introduce the overall architecture of our efficient multispectral object detector in Section~\ref{sec:efficient}. 

\subsection{Shape-Priority Early-Fusion Module}\label{sec:ShaPE}
\begin{figure*}[t]
	\includegraphics[width=\linewidth]{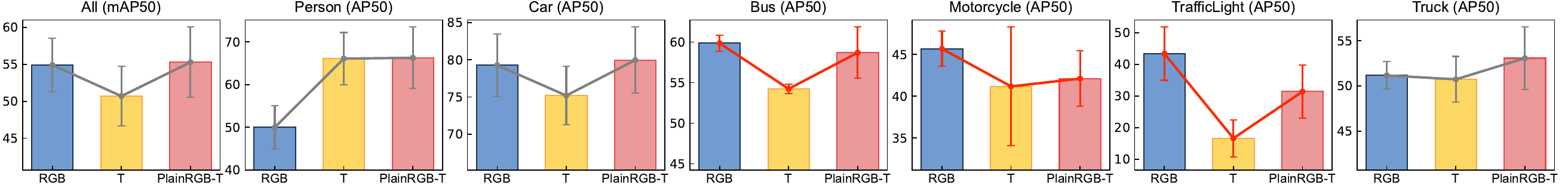}
	\caption{Pilot studies conducted on the M3FD~\cite{liu2022target} dataset. We use three detectors as baselines: RetinaNet~\cite{lin2017focal}, GFL~\cite{gfl} and YOLOv5~\cite{yolov5}. Each bar and error bar represents the mean values and standard deviation of the results obtained by these three detectors. `RGB' represents detectors that only take RGB images as inputs, while `T' represents detectors that only take thermal images as inputs. `PlainRGB-T' denotes detectors that use the plain early-fusion strategy. The `All' column illustrates the mAP50 for all classes, and the other columns illustrate the AP50 for specific classes. Red lines denote the plain RGB-T early fusion strategy obtains worse results compared to detectors that use single-modality inputs.}
	\label{fig:pilot}
	\vspace{-6mm}
\end{figure*}

\textbf{Observation.} Given a pair of RGB-T images, the plain early-fusion strategy concatenates them in the channel dimension and then feeds them into a detector. With the plain strategy, we conduct pilot studies on the M3FD~\cite{liu2022target} dataset. We first train three commonly used one-stage detectors: RetinaNet~\cite{lin2017focal}, GFL~\cite{gfl} and YOLOv5~\cite{yolov5}. Then, we compute the mean values and standard deviations of their detection results and illustrate the computed results in Fig.~\ref{fig:pilot}. Besides, we also train these detectors using single-modality images as input for comparisons. We have the following two observations. First, the plain early-fusion strategy cannot achieve consistent improvement compared with single-modality input. Second, for objects that require color to identify, such as `Traffic Light', the plain early-fusion strategy yields worse results than the RGB input.

%%%%%%%%%%%%%%%%%%%%%%%%%%% Figure: detector AP %%%%%%%%%%%%%%
%%%%%%%%%%%%%%%%%%%%%%%%%%%%%%%%%%

\textbf{Motivation.} We attribute the above phenomena to the convolutional inductive bias, namely, local connectivity and weight sharing. The process of 2D convolution involves two steps: (1) sampling across the concatenated RGB-T images using a regular grid $\mathcal{R}$; (2) summing the sampled values with weighting factor $\mathbf{W}$. The grid $\mathcal{R}$ determines both the receptive field size and dilation. For example, 
\begin{equation*}
	\mathcal{R} = \left\{(-3, -3), (-3, -2), \dots, (2, 3), (3, 3)\right\}
\end{equation*}
defines a 7$\times$7 kernel with dilation 1. For each position $\mathbf{p}_0$ on an out feature map $\mathrm{O}$, we have\\
\begin{equation}
\mathrm{O}(\mathbf{p}_0) =\sum_{\mathbf{p}_n\in\mathcal{R}}\sum_{j\in\{\rm rgb, t\}}\mathbf{W}_{j}(\mathbf{p}_n) \mathbf{I}_{j}(\mathbf{p}_0+\mathbf{p}_n),
\label{eq:2dconv}
\end{equation}
where $\mathbf{p}_n$ enumerates the positions in $\mathcal{R}$. 

\begin{figure}[t]
	\centering
	\includegraphics[width=\linewidth]{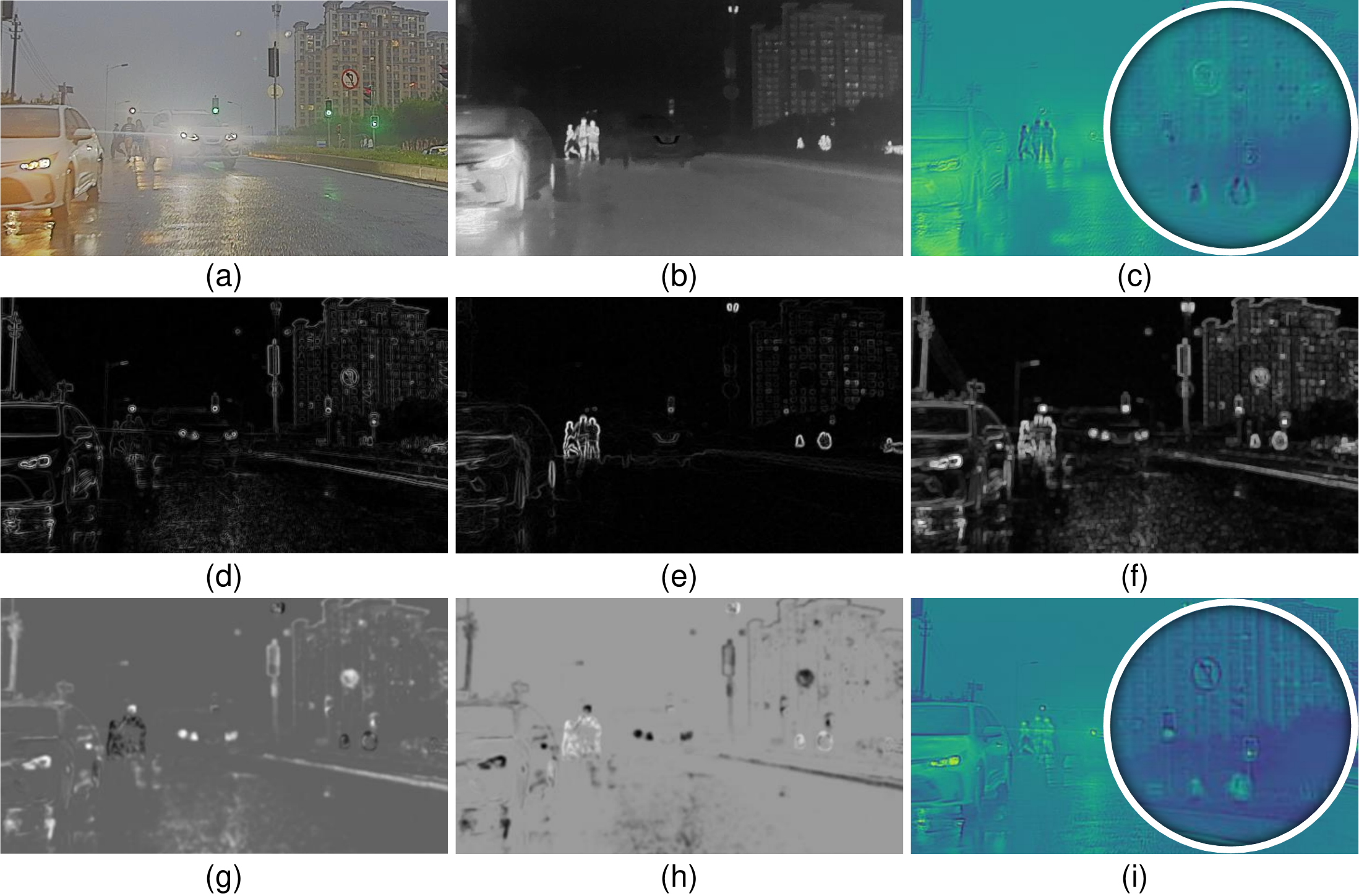}
	\vspace{-6mm}
	\caption{Illustration of fused feature map generation process for the plain early-fusion strategy and our ShaPE module. (a) RGB image. (b) Thermal image. (c) Fused feature map generated using the plain early-fusion strategy, with a close-up indicated by a white circle line. (d) and (e) are gradient images of the RGB and thermal images, respectively. (f) Boosted reference gradient image. (g) and (h) are self-gating masks of the RGB and thermal images, respectively. (i) Fused feature map generated by our ShaPE module.}
	\label{fig:earlyfusionfeature}
	\vspace{-5mm}
\end{figure}

This process indicates that the plain early-fusion strategy is a pixel-level weighting method, with weights learned from data. However, the limited receptive field of pixel-level weighting methods makes the weights difficult to determine which modality is important. This weakness may result in valuable information from one modality being suppressed by another. As an example, Fig.~\ref{fig:earlyfusionfeature}~(c) depicts the feature map generated from the RGB-T images of Fig.~\ref{fig:earlyfusionfeature}~(a) and (b) using the plain early-fusion strategy. It is observed from the close-up that the `Traffic Light' in the fused feature map doesn't preserve the significant information of the RGB image. 

The straightforward solutions to this weakness are: (1) enlarging the receptive field by using a larger kernel or more convolutional layers so that the model can judge the modality importance based on a broader range of contexts, or (2) increasing the number of convolutional kernels so that the model can learn more representations. However, these solutions increase memory costs and computational burden, making them unfriendly to edge devices.

\textbf{ShaPE Module.} We realize that shape is an inherent attribute of an object. Any visible objects in RGB and thermal images have consistent shapes. Thus, we consider the salience of shape as a modifying factor to adaptively determine the modality importance, and design the \underline{sha}pe-\underline{p}riority  \underline{e}arly-fusion (ShaPE) module. In the ShaPE module, the RGB and thermal images are modified by self-gating masks. In this context, Eq.~\eqref{eq:2dconv} becomes:\\
\begin{equation}
\mathrm{O}(\mathbf{p}_0) =\sum_{\mathbf{p}_n\in\mathcal{R}}\sum_{j\in\{\rm rgb, t\}}\mathbf{W}_ j(\mathbf{p}_n)\mathbf{M}_j(\mathbf{p}_0+\mathbf{p}_n)\mathbf{I}_{j}(\mathbf{p}_0+\mathbf{p}_n),
\label{eq:shape}
\end{equation}
where $\mathbf{M}_{\rm rgb}$ and $\mathbf{M}_{\rm t}$ denote the self-gating masks of RGB and thermal images, respectively.

In the following, we describe the generation process of self-gating masks $\mathbf{M}_{\rm rgb}$ and $\mathbf{M}_{\rm t}$. Since our ShaPE module focuses on the shapes of objects and structural contributions of different modalities to the fused features, we employ the gradients and structural similarities in our method. For easy understanding, we visualize some important intermediate results in Fig.~\ref{fig:earlyfusionfeature}. Given the RGB-T images as shown in Fig.~\ref{fig:earlyfusionfeature}~(a) and (b), we compute their gradients\\
\begin{equation*}
\begin{aligned}
	\nabla \mathbf{I}_{\rm rgb}(\mathbf{p}_0) &= \sqrt{(\nabla_x \mathbf{I}_{\rm rgb}(\mathbf{p}_0))^2 + (\nabla_y \mathbf{I}_{\rm rgb}(\mathbf{p}_0))^2},\\
	\nabla \mathbf{I}_{\rm t}(\mathbf{p}_0) &= \sqrt{(\nabla_x \mathbf{I}_{\rm t}(\mathbf{p}_0))^2 + (\nabla_y \mathbf{I}_{\rm t}(\mathbf{p}_0))^2}, 
\end{aligned}
\end{equation*}
as shown in Fig.~\ref{fig:earlyfusionfeature}~(d) and (e). We then generate the union gradient as the reference using\\
\begin{equation*}
\nabla \mathbf{I}'_{\rm ref}(\mathbf{p}_0) = \max(\nabla \mathbf{I}_{\rm rgb}(\mathbf{p}_0), \nabla \mathbf{I}_{\rm t}(\mathbf{p}_0)).
\end{equation*}
We further use max-pooling within a 3$\times$3 neighborhood $\mathcal{R}'$ to boost the reference gradient, which is written as\\
\begin{equation*}
	\nabla\mathbf{I}_{\rm ref}(\mathbf{p}_0) = \max_{\mathbf{p}_n\in\mathcal{R}'}\nabla \mathbf{I}'_{\rm ref}(\mathbf{p}_0+\mathbf{p}_n),
\end{equation*}
as shown in Fig.~\ref{fig:earlyfusionfeature}~(f).

%%%%%%%%%%%%%%%%%%%%%%%%%%%%%%%%%%%%%%%%% Figure. EarlyFusion Feature%%%%%%%%%%%%%%
%%%%%%%%%%%%%%%%%%%%%%%%%%%%%%%%%%%%%%%%%%%%%%%%%%%%%%%%%%%%%%%%%%%

To determine the structural contributions of each modality to the fused features, we compute the structural similarities between single-modality gradient images \{$\nabla\mathbf{I}_{\rm rgb}$, $\nabla\mathbf{I}_{\rm t}$\} and the reference gradient image $\nabla\mathbf{I}_{\rm ref}$. Inspired by~\cite{SSIM}, for each patch $\mathcal{R}$, we compute three fundamental properties: the means \{$\mu_{\rm rgb}, \mu_{\rm t}, \mu_{\rm ref}$\}, the standard deviations \{$\sigma_{\rm rgb}, \sigma_{\rm t}, \sigma_{\rm ref}$\}, and the covariances \{$\sigma_{({\rm rgb, ref})}$, $\sigma_{({\rm t, ref})}$\} between the single-modality gradient images and the reference gradient images. In this context, we generate the self-gating masks:\\
\begin{equation*}
\begin{aligned}
	&\mathbf{M}_{\rm rgb}' = \frac{(2\mu_{\rm rgb}\mu_{\rm ref}+\xi_1)(2\sigma_{\rm (rgb, ref)}+\xi_2)}{(\mu_{\rm rgb}^2+\mu_{\rm ref}^2+\xi_1)(\sigma_{\rm rgb}^2+\sigma_{\rm ref}^2+\xi_2)},\\
	&\mathbf{M}_{\rm t}' = \frac{(2\mu_{\rm t}\mu_{\rm ref}+\xi_1)(2\sigma_{\rm (t, ref)}+\xi_2)}{(\mu_{\rm t}^2+\mu_{\rm ref}^2+\xi_1)(\sigma_{\rm t}^2+\sigma_{\rm ref}^2+\xi_2)},
\end{aligned}
\end{equation*}
where $\xi_1=(k_1L)^2$ and $\xi_2=(k_2L)^2$ are used to prevent instability. $L$ is
the dynamic range of the gradient images, $k_1 = \text{0.01}$, and $k_2 = \text{0.03}$.

Since the ranges of both $\mathbf{M}_{\rm rgb}'$ and $\mathbf{M}_{\rm t}'$ are $[-\text{1}, \text{1}]$, we then normalize the self-gating masks and obtain\\
\begin{equation}
\mathbf{M}_{\rm rgb} = \frac{\exp(\mathbf{M}'_{\rm rgb})}{\sum\limits_{j\in\{\rm rgb, t\}}\exp(\mathbf{M}'_{j})},\; \mathbf{M}_{\rm t} = \frac{\exp(\mathbf{M}'_{\rm t})}{\sum\limits_{j\in\{\rm rgb, t\}}\exp(\mathbf{M}'_{j})},
\end{equation}
as shown in Fig.~\ref{fig:earlyfusionfeature}~(g) and (h). According to Eq.~\eqref{eq:shape}, we can finally generate the fused feature map as shown in Fig.~\ref{fig:earlyfusionfeature}~(i).

\subsection{Weakly Supervised Learning Method}\label{sec:weakly}
\begin{figure}[t]
\centering
\includegraphics[width=\linewidth]{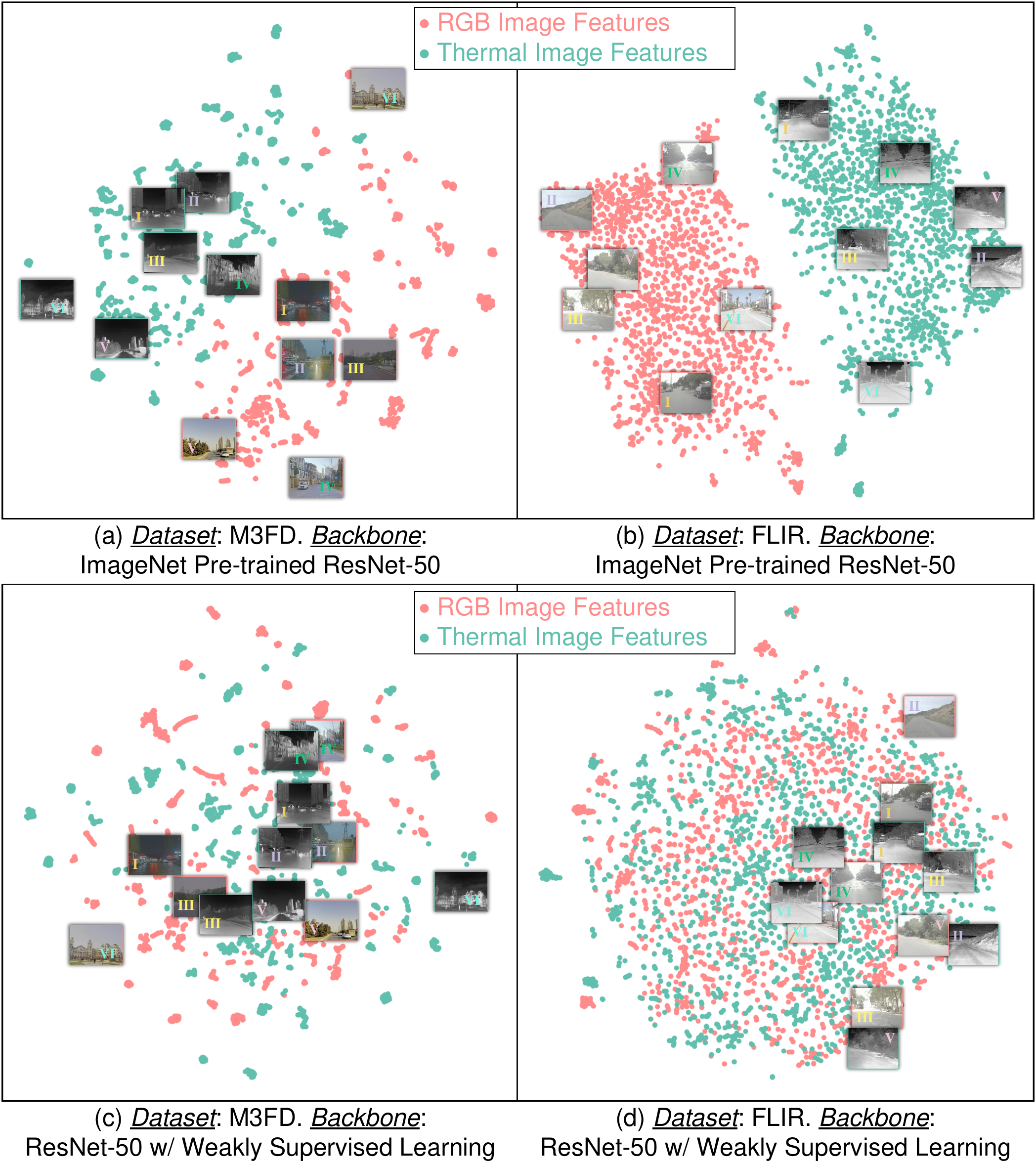}
\caption{T-SNE visualization of RGB and thermal image features. (a) and (b) visualize the image features of the M3FD~\cite{liu2022target} and FLIR~\cite{Flir} datasets using the ImageNet pre-trained ResNet-50 backbone network. (c) and (d) visualize the image features of the same datasets using the ResNet-50 trained with our weakly supervised learning method. Additionally, we present corresponding images of six pairs of feature points.}
\label{fig:datasetTSNE}
%\vspace{-8mm}
\end{figure}

In RGB-T object detection, an unneglectable issue is the lack of pre-trained backbone networks on large-scale RGB-T datasets. This is because there are few large-scale datasets like ImageNet~\cite{ILSVRC15} and COCO~\cite{lin2014microsoft} in RGB-T image recognition fields. Previous works generally use backbone networks pre-trained on ImageNet. However, the domain gap between thermal and RGB images would cause representation distribution shifts, as illustrated in Fig.~\ref{fig:datasetTSNE}~(a) and (b). This is because the backbone network is trained solely on RGB images, but is applied to thermal images.

%%%%%%%%%%%%%%%%%%%%%%%%%%%%%Figure: t-sne%%%%%%%%%%%%%%%%
%%%%%%%%%%%%%%%%%%%%%%%%%%%%%%%%%%%%%%

\begin{figure}[t!]
	\centering
	\includegraphics[width=\linewidth]{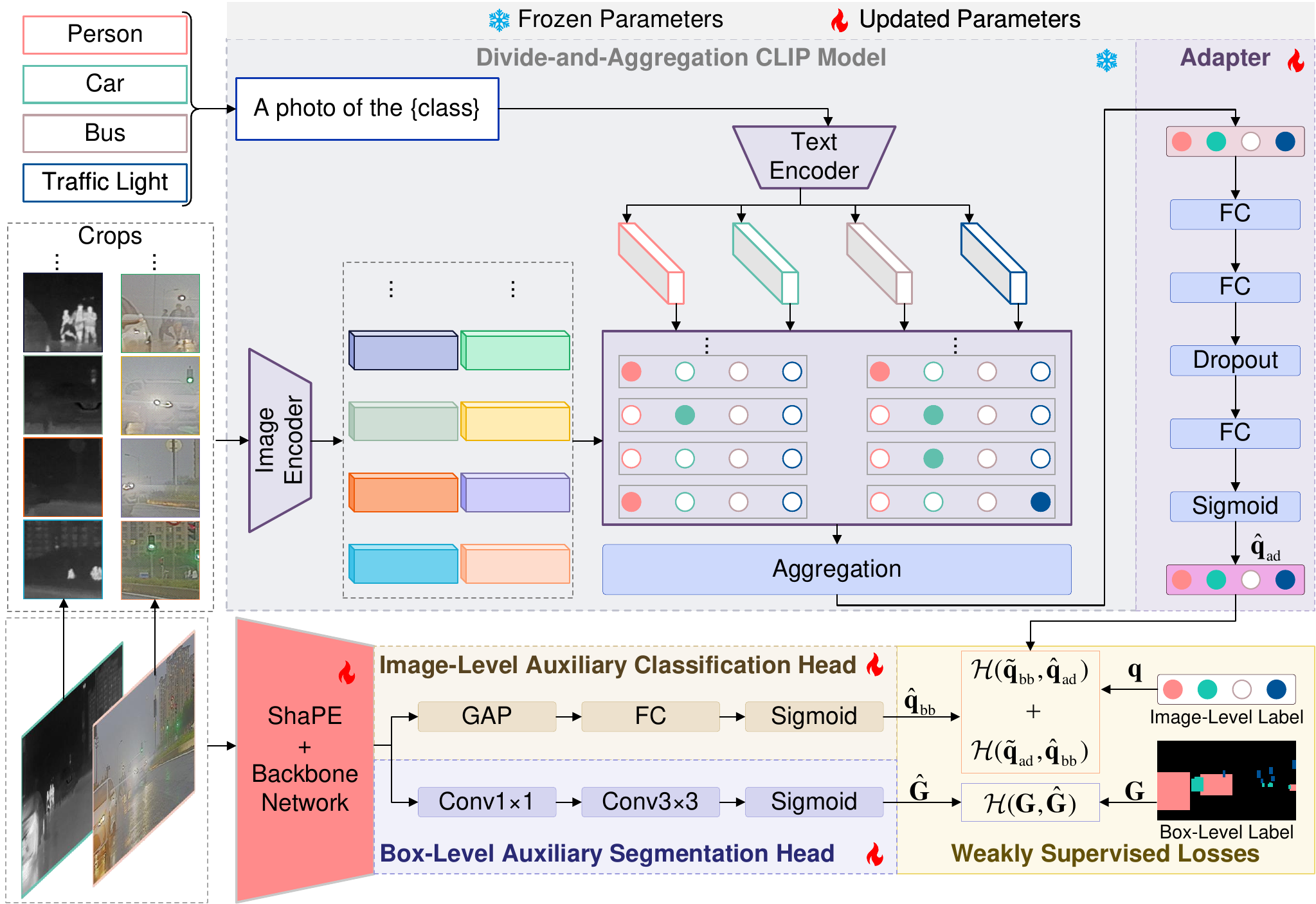}
	\caption{Illustration of the weakly supervised learning method. It consists of a divide-and-aggregation CLIP model (DA-CLIP), an adapter, a backbone, two auxiliary heads used for classification and segmentation, and weakly supervised losses. {The crops are obtained using PyTorch's function \texttt{torch.nn.functional.unfold(image, kernel\_size=224, stride=112)}. The image-level label is determined through a two-step process: 1) gather all classes present in the image according to bounding-box annotations, and 2) remove duplicated classes. Note that all modules except the DA-CLIP are updated, and only the backbone network remains in the inference phase.}}
	\label{fig:clip}
%	\vspace{-8mm}
\end{figure}

To handle this issue, we turn to the powerful Contrastive Language-Image Pre-training (CLIP)~\cite{clip} model. {It has been confirmed that CLIP can bridge domain gaps~\cite{PADCLIP, unveilingclip, textif, clipdrivenperson, unidetector}, since it is trained using a huge number of (image, text) pairs.} In this context, we feed both RGB and thermal images into the backbone network, and let it learn the representation generated by the CLIP model. Specifically, we first present a CLIP-driven image-level weakly supervised learning method. This method enables the network to recognize the classes of objects in a pair of RGB-T images while locating their coarse regions. For fine-grained localization, we then introduce a box-level weakly supervised learning method. Fig.~\ref{fig:clip} illustrates the architecture of weakly supervised learning method.

\textbf{CLIP-Driven Image-Level Weak Supervision.} To learn the CLIP model's knowledge, we construct the image-level weak supervision method. Based on three considerations, we adopt the multi-label classification task as the image-level weak supervision: (1) the CLIP model can be viewed as a classifier, (2) this auxiliary task can fully use the complementary information in the RGB-T images, and (3) by summarizing all classes and removing duplicates in an image, we can easily construct the ground-truth multi-label targets based on detection annotations.

Nevertheless, original CLIP model is only trained for recognizing a single object per image~\cite{clip} and is not suitable for multi-label classification~\cite{cdul,regionclip}. To address this issue, we introduce a Divide-and-Aggregation CLIP (DA-CLIP) model. DA-CLIP first divides input images into multiple crops. Each crop is then fed into CLIP. All predictions of these crops are finally aggregated by a max-pooling operation on each class. Considering DA-CLIP may generate inaccurate predictions, we construct a learnable adapter, which consists of three fully-connected (FC) layers, to fine-tune the result of DA-CLIP. To prevent overfitting, we add a dropout layer in the adapter. We denote the predicted probability from the adapter as $\mathbf{\hat{q}}_{\rm ad}\in\mathbb{R}^{c}$, where $c$ denotes the number of classes.

%%%%%%%%%%%%%%%%%%%%%%%%%%%CLIP%%%%%%%%%%%%
%%%%%%%%%%%%%%%%%%%%%%%%%%%%%%%%%%%%%%%%%%%%%%%%%

For the backbone network, we add an auxiliary classification head on its top. The head consists of a global average pooling (GAP) operation and one FC layer. We denote the predicted probability from the classification head as $\mathbf{\hat{q}}_{\rm bb}\in\mathbb{R}^{c}$.

We adopt the mutual learning approach~\cite{mutual} to train the backbone network and the adapter simultaneously. In this approach, an important step is that one model generates soft targets for the other model using the softmax function. However, this approach cannot be directly applied to the multi-label classification problem, since it requires the sum of predicted probabilities to be one, which is rarely satisfied in multi-label classification. To address this issue, we draw inspiration from self-training KD~\cite{kdlsr} and construct the soft targets for the adapter and backbone network as\\
\begin{equation*}
	\mathbf{\tilde{q}}_{\rm ad} = (1 - \lambda)\mathbf{q}+\lambda\mathbf{\hat{q}}_{\rm ad}, \quad \mathbf{\tilde{q}}_{\rm bb} = (1 - \lambda)\mathbf{q}+\lambda\mathbf{\hat{q}}_{\rm bb},
\end{equation*}
where $\mathbf{q}\in\mathbb{R}^{c}$ denotes a ground-truth multi-label target, and $\lambda$ denotes a balancing factor set to 0.1. In this context, we compute the binary cross-entropy (BCE) losses\\
\begin{subequations}
	\vspace{-4mm}
\begin{align}
	&\mathcal{H}(\mathbf{\tilde{q}}_{\rm ad}, \mathbf{\hat{q}}_{\rm bb})\notag\\&= -\sum_{i=1}^{c}\tilde{q}_{{\rm ad},i}\log(\hat{q}_{{\rm bb}, i})+(1-\tilde{q}_{{\rm ad},i})\log(1-\hat{q}_{{\rm bb},i}),\\
	&\mathcal{H}(\mathbf{\tilde{q}}_{\rm bb}, \mathbf{\hat{q}}_{\rm ad})\notag\\&= -\sum_{i=1}^{c}\tilde{q}_{{\rm bb},i}\log(\hat{q}_{{\rm ad}, i})+(1-\tilde{q}_{{\rm bb},i})\log(1-\hat{q}_{{\rm ad},i}).
\end{align}
\label{eq:imgloss}
\end{subequations}

%%%%%%%%%%%%%%%%%%%%%% Figure. CAM %%%%%%%%%%%%%%%%%%%%%%%%%%%%%%%%
\begin{figure}[t]
\begin{minipage}{\linewidth}
\centering
\includegraphics[width=\linewidth]{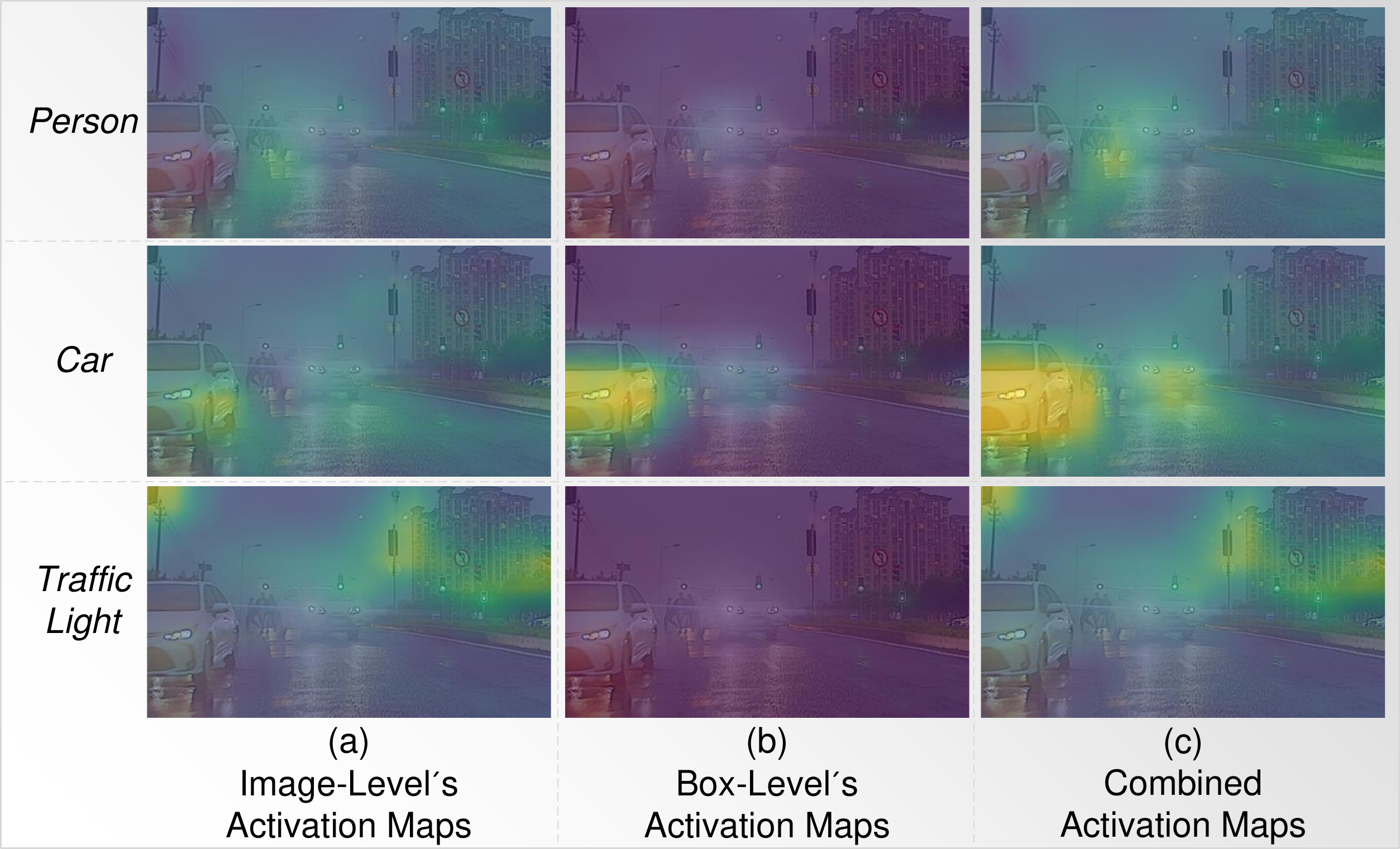}
\caption{Illustration of the class activation map (CAM) of the backbone network. Each row's triplet of images represents the CAM for a specific class, using (a) image-level auxiliary learning only, (b) box-level auxiliary learning only, and (c) both image-level and box-level auxiliary learning.}
\label{fig:fig-auxiliaryAttn}
\vspace{2mm}
\end{minipage}
\begin{minipage}{\linewidth}
\centering
\includegraphics[width=\linewidth]{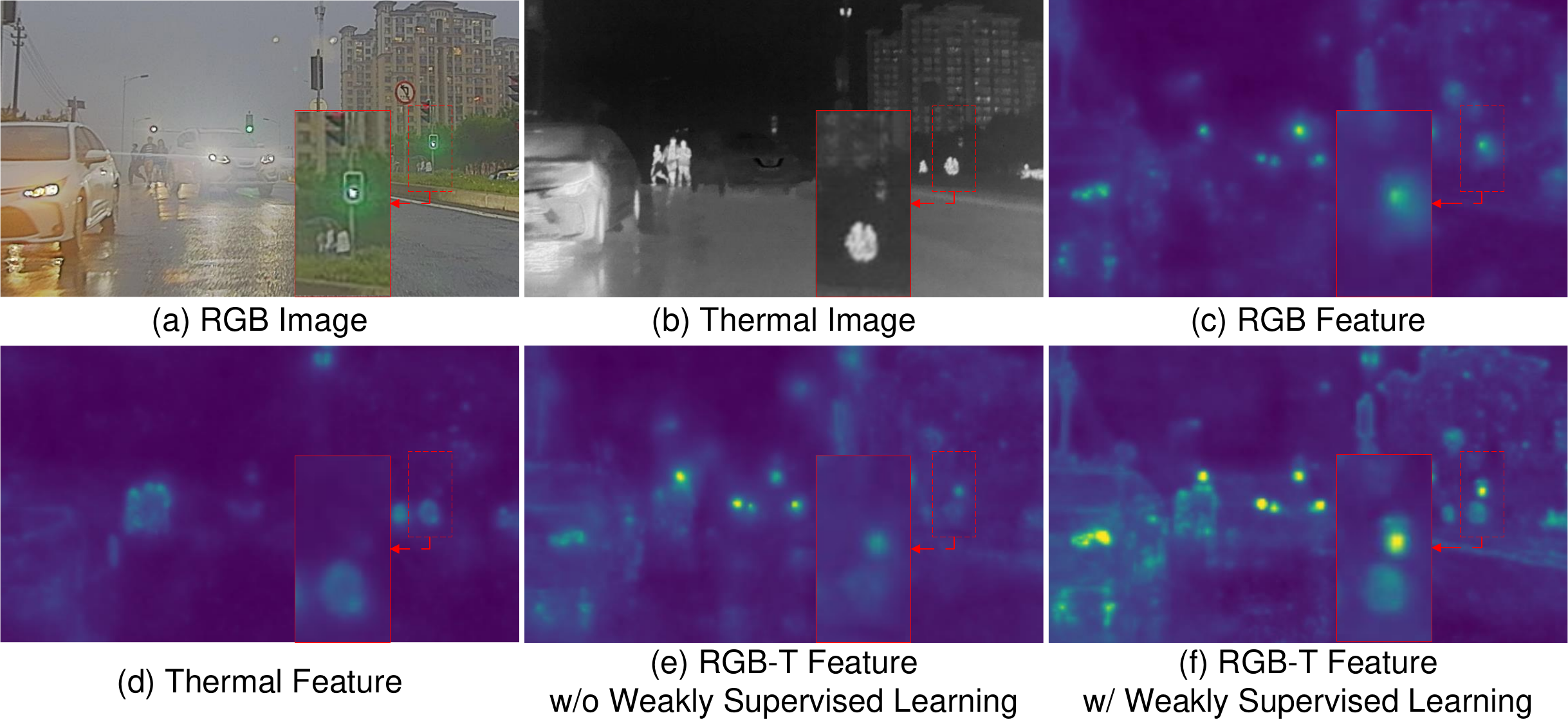}
\caption{Illustration of feature maps generated by the backbone network. (a) and (b) present the RGB and thermal images. (c) and (d) present their corresponding features map. (e) and (f) present the feature maps generated by the ResNet-50 trained without and with our weakly supervised learning, respectively. The close-up is highlighted with a red box.}
\label{fig:weaklysupfeature}
\end{minipage}
\vspace{-5mm}
\end{figure}
%%%%%%%%%%%%%%%%%%%%%%%%%%%%%%%%%%%%%%

To showcase the semantic localization effect of our CLIP-driven image-level weak supervision, we visualize the class activation map (CAM) of the backbone network in Fig.~\ref{fig:fig-auxiliaryAttn} (a). CAM is a useful tool for understanding which regions the network focuses on to predict a class. We can observe that the backbone network can coarsely localize regions of `Person', `Car', and `Traffic Light' in the image.

\textbf{Box-Level Weak Supervision.} To precisely localize the semantic regions, we introduce box-level weak supervision. The ground-truth box-level target is generated by directly filling the area within an annotation box with its corresponding class index. In this context, we add an auxiliary segmentation head on top of the backbone network to predict the target. Denoting the ground-truth box-level target mask as $\mathbf{G}$, and the predicted mask as $\mathbf{\hat{G}}$, we compute the BCE loss between them as\\
\begin{equation}
\mathcal{H}(\mathbf{G}, \mathbf{\hat{G}}) = -\sum_{n=1}^{N}G_{n}\log(\hat{G}_n)+(1-G_n)\log(1-\hat{G}_n),
\label{eq:maskloss}
\end{equation}
where $N$ denotes the number of elements in the mask.

%%%%%%%%%%%%%%%%%%%%%%%% fig: corekd%%%%%%%%%%%%
\begin{figure*}[t]
	\centering
	\includegraphics[width=0.7\linewidth]{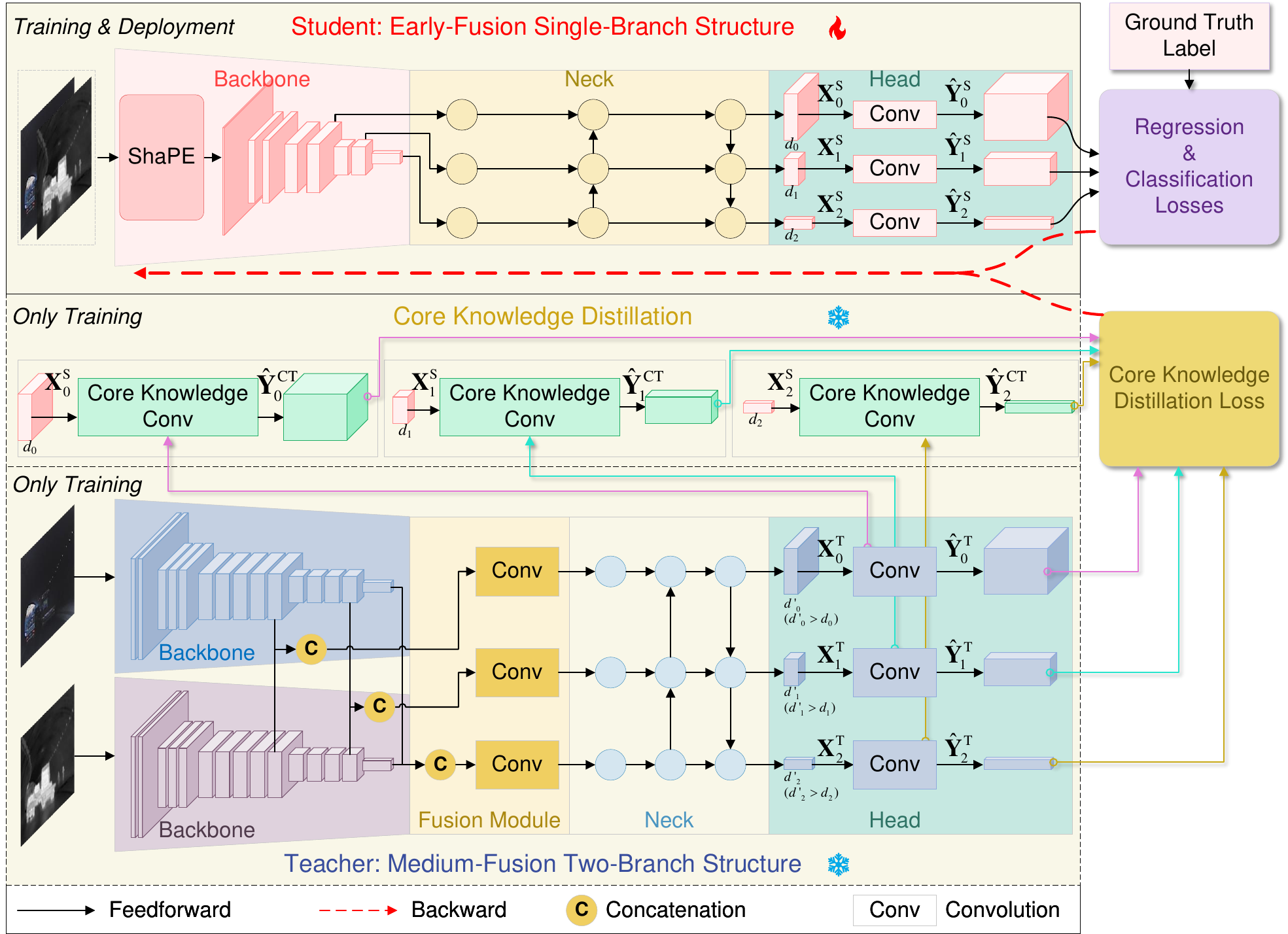}
	\caption{Illustration of the knowledge distillation technique. The student model adopts an early-fusion single-branch structure, while the teacher model adopts a medium-fusion two-branch structure. In the training phase, both the pre-trained teacher model and the core knowledge convolution module are fixed, while only the student model is updated. After training, only the student model is used for deployment. In this diagram, we use YOLOv5~\cite{yolov5} as an example, and it can be easily extended to other detectors.}
	\label{fig:kdarchitecture}
	\vspace{-6mm}
\end{figure*}
%%%%%%%%%%%%%%%%%%%%%%%%%%%%%%%%%%%%%%%%%%%%%%

We visualize attention maps of the backbone network for different classes, as shown in Fig.~\ref{fig:fig-auxiliaryAttn} (b). Using the box-level weak supervision, the backbone network can precisely localize the interest of objects, such as 'Car'. Nevertheless, it may miss some useful information in the image. Therefore, we combine the CLIP-driven image-level weak supervision and the box-level weak supervision. The results presented in Fig.~\ref{fig:fig-auxiliaryAttn} (c) show that our weakly supervised learning method can effectively allow the backbone network to localize the important semantic regions.

\textbf{Effect Validation.} When our weakly supervised learning method is employed, Fig.~\ref{fig:datasetTSNE}~(c) and (d) demonstrate that the domain gap between RGB and thermal features is reduced. This implies that the backbone network can extract information from RGB and thermal images without bias. To further illustrate this effect, we visualize the feature map generated by the ResNet-50~\cite{resnet} in Fig.~\ref{fig:weaklysupfeature}.
The generation process of these feature maps is as follows: First, we resize all features of the ResNet-50 across four stages to the same resolution as the input images. Then, we aggregate these features along the channel dimension using $\texttt{sum}(\texttt{softmax}(\mathbf{F}, \texttt{dim=0})\otimes\mathbf{F}, \texttt{dim=0})$, where $\mathbf{F}\in\mathbb{R}^{D\times H\times W}$ represents the concatenated feature. $D$, $H$, and $W$ denote its depth, height, and width, respectively. $\otimes$ denotes the element-wise production operation.

Fig.~\ref{fig:weaklysupfeature}~(a) and (b) present the RGB and thermal images in one example scene. Fig.~\ref{fig:weaklysupfeature}~(c) and (d) illustrate their corresponding feature maps. Fig.~\ref{fig:weaklysupfeature}~(e) shows the RGB-T feature map without using our weakly supervised learning method. Fig.~\ref{fig:weaklysupfeature}~(f) shows the feature map using our weakly supervised learning method. Observing Fig.~\ref{fig:weaklysupfeature}~(e), we note that the ResNet-50 tends to acquire information primarily from the RGB image. In contrast, the feature map in Fig.~\ref{fig:weaklysupfeature}~(f) demonstrates that our method enables the ResNet-50 to gather important information from both RGB and thermal images.

\subsection{Core Knowledge Distillation}\label{sec:CoreKD}
\textbf{Problem Description.} To further improve the detection accuracy of the early-fusion strategy without increasing its computational cost, we introduce the knowledge distillation technique~\cite{hinton2015distilling}. To achieve knowledge transfer, we instruct the student model to mimic intermediate features of teacher model. In this process, a primary obstacle the student model faces is the unequal number of feature channels as the teacher model. Previous works introduce convolution layers to align their feature channel numbers~\cite{fitnet, chen2017learning}, while neglecting whether the teacher's knowledge is helpful to the student. To address this issue, we propose core knowledge distillation (CoreKD).

\textbf{CoreKD Architecture.} We use YOLOv5~\cite{yolov5} as an example and illustrate the knowledge distillation architecture in Fig.~\ref{fig:kdarchitecture}. In its architecture, we use the early-fusion single-branch structure as the student model and the medium-fusion two-branch structure as the teacher model. In the student model, a pair of RGB-T images is first concatenated, then fed into different network modules, and finally converted into predicted results. In the teacher model, the RGB and thermal images are respectively fed into different backbone networks. The generated multispectral features are fused in the feature space through concatenation and convolution operations. The fused features are then fed into the subsequent network modules and converted into predicted results. The predicted results of both the student and teacher models consist of bounding boxes and class-specific confidence scores.

\textbf{CoreKD Formulation.} Since we apply the same distillation techniques to different feature pyramid levels, we only describe the technique at one level and omit the subscript for simplicity. In the head modules of Fig.~\ref{fig:kdarchitecture}, we denote the input features of the student and teacher models as $\mathbf{X}^{\rm S}$ and $\mathbf{X}^{\rm T}$, respectively. Feature distillation typically transfers the teacher's knowledge to the student by minimizing the loss~\cite{fitnet}\\
\begin{equation}
\mathcal{L}''_{\rm feat} = ||\mathcal{A}(\mathbf{X}^{\rm S}) - \mathbf{X}^{\rm T}||_2^2,
\label{eq:featureloss_naive}
\end{equation}
where $\mathcal{A}$ denotes an adaptation layer used to match the channel dimensions between the student and teacher features. Previous works usually use a convolution layer as the adaptation layer~\cite{fitnet, chen2017learning}. This approach aims to make $\mathcal{A}(\mathbf{X}^{\rm S})$ learn all information in the teacher feature $\mathbf{X}^{\rm T}$. However, they neglect whether all the information in $\mathbf{X}^{\rm T}$ is beneficial for downstream tasks, including classification and regression.

\begin{figure}
	\includegraphics[width=\linewidth]{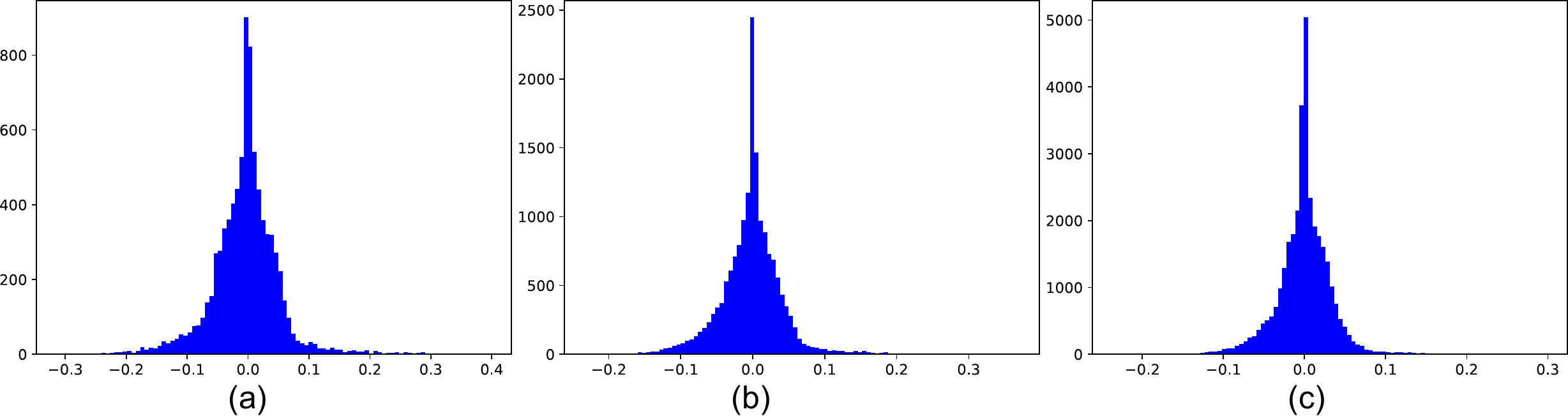}
	\caption{Weighting factor histograms of the teacher's head module in Fig.~\ref{fig:kdarchitecture}. (a), (b), and (c) correspond to the level-0, level-1, and level-2 convolution weighting factor histograms, respectively.}
	\label{fig:weightHistogram}
	\vspace{-6mm}
\end{figure}

To address this problem, we revisit the structure of head module in the teacher model. As shown in Fig.~\ref{fig:kdarchitecture}, the official implementation of YOLOv5 uses a `$1\times1\;\texttt{Conv}$' layer to output the predicted results\\
\begin{equation*}
\mathbf{\hat{Y}}^{\rm T} = \texttt{Conv}(\mathbf{X}^{\rm T}; \mathbf{W}^{\rm T}),
\end{equation*}
where $\mathbf{W}^{\rm T}$ denotes the weighting factor in the teacher's head module. According to the 2D convolution formulation in Eq.~\eqref{eq:2dconv}, we can infer that the weighting factor $\mathbf{W}^{\rm T}$ reflects the importance of a channel map in $\mathbf{X}^{\rm T}$ for the downstream feature. We visualize the histogram of $\mathbf{W}^{\rm T}$ in Fig.~\ref{fig:weightHistogram}. It is evident that most of the values in $\mathbf{W}^{\rm T}$ approximate 0. This implies that only a few feature representations in $\mathbf{X}^{\rm T}$ are important for the downstream tasks. We call these important feature representations the core knowledge in teacher model.

To learn this core knowledge, we modify the feature loss Eq.~\eqref{eq:featureloss_naive} into\\
\begin{equation}
\mathcal{L}'_{\rm feat} = ||\texttt{Conv}(\mathcal{A}(\mathbf{X}^{\rm S}); \mathbf{W}^{\rm T}) - \texttt{Conv}(\mathbf{X}^{\rm T}; \mathbf{W}^{\rm T}))||_2^2.
\label{eq:featureloss_a}
\end{equation}
This modification ensures that $\mathcal{A}(\mathbf{X}^{\text{S}})$ and $\mathbf{X}^{\text{T}}$ are projected into an identical space constructed by $\mathbf{W}^{\text{T}}$, and that the projected features are close to each other. Furthermore, to avoid introducing the adaption layer $\mathcal{A}$, we construct a core knowledge convolution (Core Knowledge Conv) operator by sampling the weighting factor $\mathbf{W}^{\rm T}$. We denote the sampling process as $\mathcal{S}(\cdot)$. In the process, we first obtain the channel dimension $d$ of the student feature $\mathbf{X}^{\rm S}$, then sample the top-$d$ values along the `\texttt{in\_channel}' axis from $\mathbf{W}^{\rm T}$ based on their absolute values. Finally, we obtain the sampled weighting factor $\mathcal{S}(\mathbf{W}^{\rm T})$. In this context, we rewrite the feature loss given in Eq.~\eqref{eq:featureloss_a} as\\
\begin{equation}
\begin{aligned}
\mathcal{L}_{\rm feat} &=||\mathbf{\hat{Y}}^{\rm CT} - \mathbf{\hat{Y}}^{\rm T}||_2^2\\ &=||\texttt{Conv}(\mathbf{X}^{\rm S}; \mathcal{S}(\mathbf{W}^{\rm T})) - \texttt{Conv}(\mathbf{X}^{\rm T}; \mathbf{W}^{\rm T}))||_2^2,
\end{aligned}
\label{eq:featureloss_corekd}
\end{equation}
where $\mathbf{\hat{Y}}^{\rm CT}$ denotes the output of core knowledge convolution. When using this feature loss, we keep the weighting factor $\mathbf{W}^{\rm T}$ fixed and only compute the gradient with respect to the student feature $\mathbf{X}^{\rm S}$.
%%%%%%%%%%%%%%%% fig: weight histogram%%%%%%%%%%%%%%%%
%%%%%%%%%%%%%%%%%%%%%%%%%%%%%%%%%%%%%%%%%%%%%%%%%%%%

{
\textbf{Mathematical Foundation of CoreKD.}
We first explain the mathematical foundation of traditional feature distillation and analyze its weaknesses. Then, we introduce the mathematical foundation of our CoreKD. Finally, we compare the results of our CoreKD with those of the traditional one.

Traditional feature distillation uses a convolution layer to align feature channel numbers between student and teacher models. We denote the convolution layer as a function $\mathcal{A}(\cdot)$ in the Eq.~\eqref{eq:featureloss_naive}.  Next, we denote the weighting factor of $\mathcal{A}(\cdot)$ as $\mathbf{A}\in\mathbb{R}^{d'\times d}$. This function is used to convert an input feature $\mathbf{X}^{\rm S}\in\mathbb{R}^{h\times w \times d}$ into an output feature $\mathbf{Z}\in\mathbb{R}^{h\times w \times d'}$, i.e., $\mathbf{Z} = \mathcal{A}(\mathbf{X}^{\rm S})$. We denote the vector at an arbitrary spatial location of $\mathbf{X}^{\rm S}$  as $\mathbf{x}^{\rm S}\in\mathbb{R}^{d\times1}$, the \textit{i}th row vector in the weighting factor $\mathbf{A}$ as $\mathbf{a}_i\in\mathbb{R}^{1\times d}$, and the corresponding value in $\mathbf{Z}$ as $z_i\in\mathbb{R}^{1\times1}$. The mathematical relation between $\mathbf{x}^{\rm S}$, $\mathbf{a}_i$, and $z_i$ can be represented as
\begin{equation}
	z_i = \texttt{Conv}(\mathbf{x}^{\rm S};\mathbf{a}_i) = \mathbf{a}_i\cdot\mathbf{x}^{\rm S},
\end{equation}
where the operator `$\cdot$' indicates a dot product. The dot product computation can be viewed as the projection of the vector $\mathbf{x}^{\rm S}$ onto the vector $\mathbf{a}_i$, as shown in Fig.~\ref{fig:projection}~(a). We can infer that the generation of $z_i$ is related to the $\mathbf{a}_i$ and $\mathbf{x}^{\rm S}$ but has no relation to the teacher's features. This implies that the traditional feature distillation merely focuses on enforcing the student to learn all information from the teacher without considering whether the teacher's features are beneficial to downstream tasks.

On the contrary, our CoreKD uses the weighting factor $\mathbf{W}^{\rm T}$ of the teacher model to align feature channel numbers between the student and teacher models, as shown in Eq.~\eqref{eq:featureloss_corekd}. We denote the \textit{i}th row vector of $\mathbf{W}^{\rm T}$ as $\mathbf{w}^{\rm T}$ and the vector at an arbitrary spatial location of $\mathbf{X}^{\rm T}$ as $\mathbf{x}^{\rm T}$. Then we can write the \textit{i}th loss value of $\mathcal{L}_{\rm feat}$ as
\begin{equation}
\begin{aligned}
{\ell}^i_{\rm feat} &= \texttt{Conv}(\mathbf{x}^{\rm S};\mathcal{S}(\mathbf{w}_i^{\rm T}))-\texttt{Conv}(\mathbf{x}^{\rm T};\mathbf{w}_i^{\rm T})\\
&=\mathcal{S}(\mathbf{w}^{\rm T}_i)\cdot\mathbf{x}^{\rm S} - \mathbf{w}^{\rm T}_i\cdot \mathbf{x}^{\rm T}.
\end{aligned}
\end{equation}
This loss value calculation process is illustrated in Fig.~\ref{fig:projection}~(b). From the above analyses, we have two key observations: 1) our CoreKD projects the student and teacher features into an identical space constructed by $\mathbf{W}^{\rm T}$, and 2) our CoreKD does not enforce the student feature to be the same as the teacher feature but rather focuses on minimizing the projected distances. Since the values within the weighting factor $\mathbf{W}^{\rm T}$ reflect the importance of teacher features, our CoreKD enables the student model to learn beneficial information for downstream tasks from the teacher model.

\begin{figure}
\centering
\includegraphics[width=\linewidth]{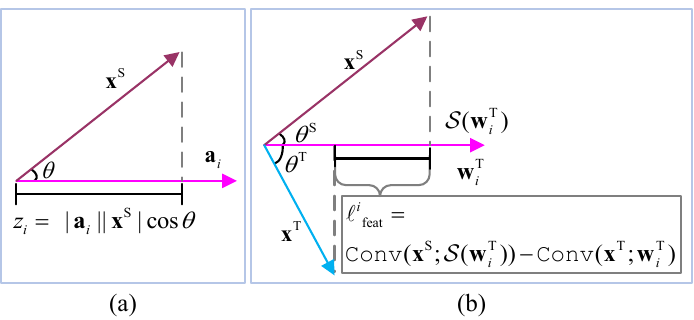}
\caption{Schematic diagram of the mathematical foundation of feature distillation: (a) Convolution operation in the traditional feature distillation; (b) The loss calculation process in our CoreKD.}
\label{fig:projection}
\vspace{-5mm}
\end{figure}

Since the experimental results involve the introduction of both datasets and implementation details, we arrange the comparison results of our CoreKD with the traditional feature distillation in the experimental section. For details, please refer to Section~\ref{sec:variants}.
}

\subsection{Loss Function}
Our efficient multispectral early-fusion (EME) single-branch model is trained using all the losses described above. The total loss is\\
\begin{equation}
	\mathcal{L}_{\rm total} = \mathcal{L}_{\rm cls} + \mathcal{L}_{\rm reg} + \mathcal{L}_{\rm weak} + \mathcal{L}_{\rm feat},
	\label{eq:totalloss}
\end{equation}
where $\mathcal{L}_{\rm cls}$ and $\mathcal{L}_{\rm reg}$ represent the classification and regression losses defined by a detector~\cite{lin2017focal, gfl, yolov5}, respectively. $\mathcal{L}_{\rm weak}$ is the summation of weakly supervised losses defined in Eq.~\eqref{eq:imgloss} and Eq.~\eqref{eq:maskloss}:\\
\begin{equation*}
	\mathcal{L}_{\rm weak} = \mathcal{H}(\mathbf{\tilde{q}}_{\rm ad}, \mathbf{\hat{q}}_{\rm bb}) + \mathcal{H}(\mathbf{\tilde{q}}_{\rm bb}, \mathbf{\hat{q}}_{\rm ad}) + \mathcal{H}(\mathbf{G}, \mathbf{\hat{G}}).
\end{equation*}

\section{Experiments}
\subsection{Experimental Setup}
\label{sec:setup}
\textbf{Datasets.}
Our experiments are conducted on the M3FD dataset~\cite{liu2022target} and FLIR dataset~\cite{Flir}. \textbf{M3FD dataset} contains 4200 pairs of RGB and thermal images. These image pairs are well aligned. The dataset contains 6 classes of objects: `Person', `Car', `Bus', `Motorcycle', `Traffic Light', and `Truck'. Since this dataset doesn't provide unified data splits, previous works have used a random splitting approach to determine the train and validation sets~\cite{liu2022target}. However, images in this dataset are sampled from video sequences, meaning that two adjacent frames may contain identical content. In this context, the random splitting approach results in information leakage between the train and validation sets. To address this problem, we first manually divide the dataset into 73 video segments based on different scenes. Then, we collect the first 70\% of images in each video segment as the train set and the remaining images as the validation set. Finally, we obtain 2905 and 1295 pairs of RGB-T images in the train and validation sets, respectively. We name this data split `M3FD-zxSplit' and release it to the public\footnote{\url{https://github.com/XueZ-phd/Efficient-RGB-T-Early-Fusion-Detection}}. For the performance evaluation in Section~\ref{sec:mainresult}, we use this data split. When comparing with state-of-the-art approaches in Section~\ref{sec:sota}, we employ both `M3FD-zxSplit' and random splitting. Our random splitting refers to randomly selecting 80\% images as the train set and the remaining images as the validation set. \textbf{FLIR dataset} originally contains unaligned RGB-T image pairs. The work \cite{alignedFLIR} develops a data-processing approach to align these images and obtain 7381, 1056, and 2111 image pairs in the train, validation, and test sets. This dataset contains 3 classes: `Person', `Bicycle' and `Car'.

\textbf{Evaluation Metrics.} We use the standard mean Average Precision (mAP) with IoU thresholds ranging from 0.5 to 0.95 across various object scales as metrics.

\begin{figure*}[h!]
\begin{minipage}{\linewidth}
\begin{center}
\captionof{table}{{Inference efficiency and detection performance on the M3FD dataset~\cite{liu2022target}. The inference time is evaluated on an edge device: NVIDIA AGX Orin. The best results in the mAP and mAP50 columns are highlighted in bold and marked in {\color{red}\textbf{red}}, while the second best ones are underlined and marked in {\color[RGB]{0,128,0}\textbf{green}}. All detection results are obtained by running three independent experiments. The mean value and standard deviation of these results are reported.}}
\label{table:mainm3fd}
\vspace{-1mm}
\resizebox{\linewidth}{!}{
{
\begin{tabular}{l|l|c|c|c|c|c|c|c|c|c|c|c}
\toprule
& Detector        & FLOPs (↓) & Parameter (↓) & Time (↓) & mAP (↑)                                    & mAP50 (↑)                                  & Person (↑) & Car (↑)    & Bus (↑)    & Motor (↑)  & TrafficLight (↑) & Truck (↑)  \\\midrule
RGB                          & RetinaNet-Res50 & 61.893G   & 36.434M       & 0.106s   & 31.03±0.09                                 & 51.30±0.16                                 & 44.57±0.12 & 74.87±0.26 & 57.80±0.24 & 44.30±0.22 & 36.63±0.58       & 49.70±0.16 \\
Thermal                      & RetinaNet-Res50 & 61.893G   & 36.434M       & 0.106s   & 29.27±0.05                                 & 46.97±0.09                                 & 59.17±0.34 & 71.17±0.29 & 54.17±0.69 & 35.90±0.22 & 10.43±0.09       & 50.83±0.21 \\\midrule
RGB-T Medium   Fusion        & RetinaNet-Res50 & 94.611G   & 47.582M       & 0.170s   & {\color[HTML]{008000} {\underline{33.43±0.05}}}    & {\color[HTML]{FF0000} \textbf{53.63±0.05}} & 60.10±0.08 & 77.27±0.05 & 61.63±0.05 & 45.43±0.09 & 25.67±0.12       & 51.80±0.00 \\\midrule
Baseline: RGB-T Early Fusion & RetinaNet-Res50 & 62.164G   & 36.434M       & 0.110s   & 32.03±0.05                                 & 50.70±0.16                                 & 58.93±0.26 & 75.37±0.09 & 58.97±0.39 & 39.97±0.21 & 21.67±0.47       & 49.33±0.05 \\
\rowcolor[HTML]{D9D9D9} 
+ ShaPE                      & RetinaNet-Res50 & 62.218G   & 36.434M       & 0.149s   & 32.80±0.08                                 & 51.97±0.24                                 & 60.20±0.59 & 77.10±0.08 & 58.83±0.33 & 39.77±0.58 & 24.57±0.54       & 51.33±0.17 \\
\rowcolor[HTML]{D9D9D9} 
+ ShaPE + WeakSup.           & RetinaNet-Res50 & 62.218G   & 36.434M       & 0.149s   & {\color[HTML]{FF0000} \textbf{33.53±0.05}} & 52.90±0.16                                 & 59.10±0.71 & 77.07±0.41 & 62.00±0.78 & 40.97±1.67 & 24.03±0.85       & 54.30±0.71 \\
\rowcolor[HTML]{D9D9D9} 
+ ShaPE + WeakSup. + CoreKD  & RetinaNet-Res50 & 62.218G   & 36.434M       & 0.149s   & {\color[HTML]{FF0000} \textbf{33.53±0.17}} & {\color[HTML]{008000} {\underline{53.23±0.09}}}    & 61.47±0.49 & 76.40±0.08 & 59.20±0.08 & 43.10±0.22 & 25.97±0.12       & 53.17±0.19 \\\midrule
RGB                          & GFL-Res50       & 61.392G   & 32.270M       & 0.110s   & 32.23±0.12                                 & 53.10±0.16                                 & 48.67±0.12 & 77.43±0.26 & 60.27±0.60 & 43.50±0.16 & 39.07±0.94       & 49.63±0.39 \\
Thermal                      & GFL-Res50       & 61.392G   & 32.270M       & 0.110s   & 29.50±0.22                                 & 48.27±0.42                                 & 64.27±0.34 & 73.73±0.12 & 52.50±1.99 & 36.50±0.0  & 15.10±0.14       & 47.37±0.25 \\\midrule
RGB-T Medium Fusion          & GFL-Res50       & 94.110G   & 43.419M       & 0.172s   & 34.17±0.33                                 & 54.47±0.60                                 & 65.37±0.12 & 79.83±0.05 & 61.20±1.00 & 37.00±1.79 & 34.80±0.57       & 48.73±0.77 \\\midrule
Baseline: RGB-T Early Fusion & GFL-Res50       & 61.663G   & 32.271M       & 0.114s   & 33.50±0.28                                 & 52.77±0.25                                 & 64.03±0.12 & 78.20±0.16 & 53.00±1.31 & 39.27±0.26 & 30.47±0.56       & 51.57±0.54 \\
\rowcolor[HTML]{D9D9D9} 
+ ShaPE                      & GFL-Res50       & 61.718G   & 32.271M       & 0.151s   & 35.17±0.17                                 & 55.50±0.22                                 & 65.80±0.24 & 79.10±0.08 & 62.33±1.24 & 41.33±0.42 & 30.80±0.42       & 53.67±0.41 \\
\rowcolor[HTML]{D9D9D9} 
+ ShaPE + WeakSup.           & GFL-Res50       & 61.718G   & 32.271M       & 0.151s   & {\color[HTML]{008000} {\underline{35.23±0.53}}}    & {\color[HTML]{008000} {\underline{55.97±0.24}}}    & 65.73±0.49 & 79.57±0.17 & 60.50±4.02 & 43.10±1.06 & 33.73±2.01       & 53.10±1.94 \\
\rowcolor[HTML]{D9D9D9} 
+ ShaPE + WeakSup. + CoreKD  & GFL-Res50       & 61.718G   & 32.271M       & 0.151s   & {\color[HTML]{FF0000} \textbf{37.03±0.09}} & {\color[HTML]{FF0000} \textbf{57.70±0.08}} & 68.43±0.25 & 81.23±0.09 & 63.37±0.33 & 43.90±0.86 & 35.77±0.12       & 53.53±0.29\\\bottomrule
\end{tabular}}}
\end{center}
\end{minipage}
\begin{minipage}{\linewidth}
\vspace{1mm}
\includegraphics[width=\linewidth]{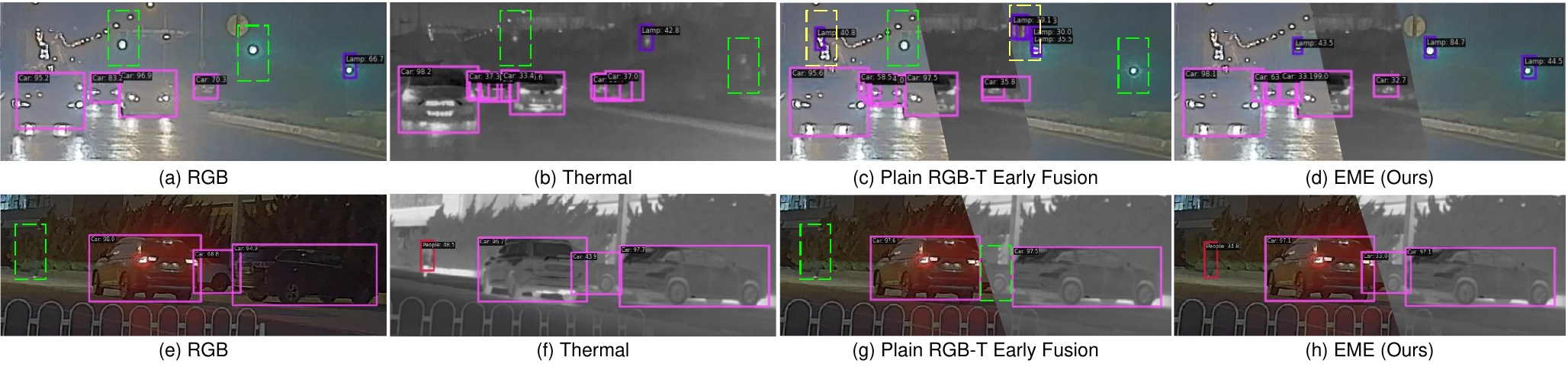}
\vspace{-6mm}
\caption{Detection results of the GFL~\cite{gfl} detector on two example scenes from the M3FD~\cite{liu2022target} dataset. (a) and (e) display results using only RGB images. (b) and (f) show results using only thermal images. (c) and (g) demonstrate results using the plain RGB-T early-fusion strategy. (d) and (h) depict results using our EME method. Solid boxes represent detection results. Green dashed boxes mark missed objects (false negatives) while yellow dashed boxes mark false positives.}
\label{fig:comparisonm3fd}
\end{minipage}
\begin{minipage}{\linewidth}
\begin{center}
\vspace{1mm}
\captionof{table}{{Inference efficiency and detection performance on the FLIR dataset~\cite{Flir}. The inference time is evaluated on an edge device: NVIDIA AGX Orin. The best results in the mAP and mAP50 columns are highlighted in bold and marked in {\color{red}\textbf{red}}, while the second best ones are underlined and marked in {\color[RGB]{0,128,0}\textbf{green}}. All detection results are obtained by running three independent experiments. The mean value and standard deviation of these results are reported.}}
\vspace{-1mm}
\label{table:flir}
\setlength{\tabcolsep}{12pt}
\resizebox{\linewidth}{!}{
{
\begin{tabular}{l|l|c|c|c|c|c|c|c|c}
\toprule
& Detector        & FLOPs (↓) & Parameter (↓) & Time (↓) & mAP (↑)                                    & mAP50 (↑)                                  & Person (↑) & Bicycle (↑)    & Car (↑)    \\\midrule
RGB                          & RetinaNet-Res50 & 61.893G   & 36.434M       & 0.106s   & 28.10±0.0                                  & 59.47±0.12                                 & 44.93±0.39 & 55.70±0.08 & 77.80±0.08 \\
Thermal                      & RetinaNet-Res50 & 61.893G   & 36.434M       & 0.106s   & 35.53±0.05                                 & 70.93±0.05                                 & 62.17±0.17 & 66.37±0.09 & 84.27±0.05 \\\midrule
RGB-T Medium Fusion          & RetinaNet-Res50 & 94.611G   & 47.582M       & 0.170s   & 38.50±0.08                                 & 71.57±0.05                                 & 61.93±0.29 & 67.60±0.14 & 85.17±0.09 \\\midrule
Baseline: RGB-T Early Fusion & RetinaNet-Res50 & 62.164G   & 36.434M       & 0.110s   & 37.47±0.05                                 & 69.57±0.05                                 & 60.70±0.22 & 63.77±0.09 & 84.37±0.05 \\
\rowcolor[HTML]{D9D9D9} 
+ ShaPE                      & RetinaNet-Res50 & 62.218G   & 36.434M       & 0.149s   & 38.70±0.14                                 & 71.60±0.22                                 & 61.40±0.54 & 68.60±0.36 & 84.70±0.22 \\
\rowcolor[HTML]{D9D9D9} 
+ ShaPE + WeakSup.           & RetinaNet-Res50 & 62.218G   & 36.434M       & 0.149s   & {\color[HTML]{008000} {\underline{38.80±0.29}}}    & {\color[HTML]{008000} {\underline{72.07±0.37}}}    & 62.50±0.96 & 68.77±0.97 & 85.03±0.21 \\
\rowcolor[HTML]{D9D9D9} 
+ ShaPE + WeakSup. + CoreKD  & RetinaNet-Res50 & 62.218G   & 36.434M       & 0.149s   & {\color[HTML]{FF0000} \textbf{38.83±0.17}} & {\color[HTML]{FF0000} \textbf{72.23±0.31}} & 62.27±0.26 & 69.23±0.66 & 85.10±0.22 \\\midrule
RGB                          & GFL-Res50       & 61.392G   & 32.270M       & 0.110s   & 31.73±0.12                                 & 63.77±0.05                                 & 51.70±0.08 & 57.87±0.25 & 81.77±0.05 \\
Thermal                      & GFL-Res50       & 61.392G   & 32.270M       & 0.110s   & 42.40±0.14                                 & 75.07±0.05                                 & 69.80±0.16 & 68.47±0.24 & 86.93±0.05 \\\midrule
RGB-T Medium Fusion          & GFL-Res50       & 94.110G   & 43.419M       & 0.172s   & 42.60±0.08                                 & 76.07±0.21                                 & 70.07±0.19 & 70.57±0.41 & 87.63±0.05 \\\midrule
Baseline: RGB-T Early Fusion & GFL-Res50       & 61.663G   & 32.271M       & 0.114s   & 41.90±0.22                                 & 74.77±0.17                                 & 69.70±0.33 & 67.73±0.26 & 87.00±0.00 \\
\rowcolor[HTML]{D9D9D9} 
+ ShaPE                      & GFL-Res50       & 61.718G   & 32.271M       & 0.151s   & 42.40±0.16                                 & 75.77±0.17                                 & 69.97±0.21 & 70.13±0.37 & 87.23±0.09 \\
\rowcolor[HTML]{D9D9D9} 
+ ShaPE + WeakSup.           & GFL-Res50       & 61.718G   & 32.271M       & 0.151s   & {\color[HTML]{008000} {\underline{42.93±0.24}}}    & {\color[HTML]{008000} {\underline{76.90±0.22}}}    & 71.30±0.14 & 71.40±0.50 & 87.97±0.09 \\
\rowcolor[HTML]{D9D9D9} 
+ ShaPE + WeakSup. + CoreKD  & GFL-Res50       & 61.718G   & 32.271M       & 0.151s   & {\color[HTML]{FF0000} \textbf{44.00±0.00}} & {\color[HTML]{FF0000} \textbf{78.17±0.05}} & 73.03±0.05 & 72.63±0.17 & 88.80±0.00\\\bottomrule
\end{tabular}
}}
\end{center}
\end{minipage}
\begin{minipage}{\linewidth}
\vspace{1mm}
\includegraphics[width=\linewidth]{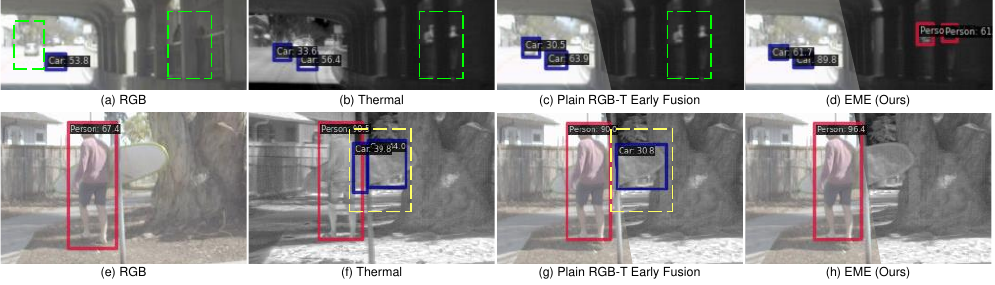}
\vspace{-6mm}
\caption{Detection results of the GFL~\cite{gfl} detector on two example scenes from the FLIR~\cite{Flir} dataset. (a) and (e) display results using only RGB images. (b) and (f) show results using only thermal images. (c) and (g) demonstrate results using the plain RGB-T early-fusion strategy. (d) and (h) depict results using our EME method. Solid boxes represent detection results. Green dashed boxes mark missed objects (false negatives) while yellow dashed boxes mark false positives.}
\label{fig:comparisonflir}
\end{minipage}
\vspace{-6mm}
\end{figure*}

\textbf{Inference Efficiency Evaluations.} We assess the inference efficiency of our method (Python implementation) on the edge device NVIDIA AGX Orin with 64GB memory. We also evaluate the complexity of our method using FLOPs and the number of parameters. All results are presented in Tables~\ref{table:mainm3fd} and \ref{table:flir}.

\textbf{Implementation Details.}
We incorporate our three key modules into commonly-used one-stage detectors, including RetinaNet~\cite{lin2017focal}, GFL~\cite{gfl}, and YOLOv5~\cite{yolov5}. For RetinaNet and GFL, we adopt the implementations in MMDetection toolbox~\cite{mmdetection}. For YOLOv5, we use its official implemtation~\cite{yolov5}.

{In the early-fusion strategy based on RetinaNet~\cite{lin2017focal} and GFL~\cite{gfl} detectors, we use ResNet-50~\cite{resnet} as the backbone network. For a fair comparison, we use the same backbone network in the medium-fusion strategy. Notably, in the CoreKD technique, the teacher model utilizes ResNet-101~\cite{resnet} as the backbone network. For both strategies, we train for 12 epochs using the SGD optimizer with a batch size of 4. The initial learning rate is set to 0.01 and is decayed by 0.1 at epochs 8 and 11. Random horizontal flipping is employed as a data augmentation technique.
	
For the early-fusion strategy based on YOLOv5, we use YOLOv5-small as the baseline detector, and we use YOLOv5-large to construct a teacher model in the CoreKD technique. For both strategies, we train for 36 epochs with a batch size of 16. We keep all other hyperparameters consistent with the official settings of the YOLOv5 repository~\cite{yolov5}.

To standardize data for RetinaNet and GFL detectors, we calculate the mean value and standard deviation of RGB and thermal images for the M3FD and FLIR datasets. All experiments use the 640 $\times$ 512 image resolution. For the M3FD dataset, we obtain mean$_{\rm{rgb}}$ = [128.2, 129.3, 125.3], std$_{\rm{rgb}}$ = [49.1, 50.2, 53.5], mean$_{\rm{t}}$ = [84.1, 84.1, 84.1], and std$_{\rm{t}}$ = [50.6, 50.6, 50.6]. For the FLIR dataset, we obtain mean$_{\rm{rgb}}$ = [149.4, 148.7, 141.7], std$_{\rm{rgb}}$ = [49.3, 52.8, 59.0], mean$_{\rm{t}}$ = [135.7, 135.7, 135.7], and std$_{\rm{t}}$ = [63.6, 63.6, 63.6]. For the YOLOv5 detector, we normalize both RGB and thermal images to the range of [0, 1] following its official implementations.}

%%%%%%%%%%%%%%%%%%%%%%%%%%%%%%%%%%%%%%% Main Results on M3FD%%%%%%%%%%%%%%%%%%
%%%%%%%%%%%%%%%%%%%%%%%%%%%%%%%%%%%%%%%%%%%%%%%%%%%%%%%%%%%%%%%%%%%%%%

\subsection{Performance Evaluation of Proposed Modules}\label{sec:mainresult}
Table~\ref{table:mainm3fd} and Table~\ref{table:flir} present the performance of our method on the M3FD~\cite{liu2022target} and FLIR~\cite{Flir} datasets. Key observations include: (1) the medium-fusion strategy adds more parameters and FLOPs compared to the early-fusion strategy; (2) the medium-fusion strategy achieves better performance compared to single-modality inputs, whereas the plain early-fusion strategy does not consistently improve performance; (3) our EME method, incorporating the ShaPE module, weakly supervised learning, and CoreKD techniques into the plain early-fusion strategy, achieves significant performance improvement without significantly increasing parameters and FLOPs; (4) the inference time of our EME method is longer than that of the baseline method, since the structural similarity computation process has not been optimized when calculating the self-gating mask; (5) our EME method can outperform the medium-fusion strategy in both performance and efficiency to some extent; {(6) Both architectures: ``Baseline + ShaPE + WeakSup." and ``Baseline + ShaPE + WeakSup. + CoreKD" have the same FLOPs, parameters, and inference time as ``Baseline + ShaPE". This is because both the weakly supervised learning method and CoreKD are removed in the inference phase, while only the ShaPE module is retained.}

Fig.~\ref{fig:comparisonm3fd} and Fig.~\ref{fig:comparisonflir} present visualization results for two example scenes from M3FD~\cite{liu2022target} and FLIR~\cite{Flir} datasets, respectively. We observe that false positives or false negatives in the single-modality results may affect the plain early-fusion strategy. For instance, the person missed in Fig.~\ref{fig:comparisonm3fd}~(e) is also absent in Fig.~\ref{fig:comparisonm3fd}~(g), despite being detected in Fig.~\ref{fig:comparisonm3fd}~(f). Moreover, false positives in Fig.~\ref{fig:comparisonflir}~(f) affect the detection results of plain early-fusion, as shown in Fig.~\ref{fig:comparisonflir}~(g). These phenomena confirm that the problem of information interference is a key obstacle to performance in the plain early-fusion strategy. Clearly, our EME effectively alleviates this problem.

{
\subsection{Performance Evaluation of Variants}\label{sec:variants}
\textbf{Feature Distillation.} We compare the results of traditional feature distillation with those of our CoreKD. Table~\ref{table:comparisonKD_m3fd} and Table~\ref{table:comparisonKD_flir} present comparison results on the M3FD and FLIR datasets, respectively. For comprehensive comparisons, we adopt RetinaNet-Res50 and GFL-Res50 as baseline detectors in these two tables. From the results, we can observe that our CoreKD consistently achieves superior performance compared to traditional feature distillation. For example, our CoreKD (72.23\%) obtains a 1.26\% mAP50 absolute gain over the traditional one (70.97\%) when using RetinaNet-Res50 on the FLIR dataset.

\begin{table}[t]
{
\caption{Comparison of traditional feature distillation with our CoreKD on the M3FD dataset~\cite{liu2022target}.}
\vspace{-1mm}
\label{table:comparisonKD_m3fd}
\setlength{\tabcolsep}{3pt}
\resizebox{\linewidth}{!}{
\begin{tabular}{l|l|l|l}
\toprule
Method & Detector        & mAP (↑)      & mAP50 (↑)    \\\midrule
\begin{tabular}[c]{@{}l@{}}Baseline:\\ Plain RGB-T Early Fusion\end{tabular}           & RetinaNet-Res50 & 32.03±0.05 & 50.70±0.16 \\\midrule
\begin{tabular}[c]{@{}l@{}}Traditional Feature Distill+\\ Baseline+ShaPE+WeakSup.\end{tabular} & RetinaNet-Res50 & 32.17±0.12 & 52.43±0.05 \\\midrule
\rowcolor[HTML]{D9D9D9} 
\begin{tabular}[c]{@{}l@{}}CoreKD+\\ Baseline+ShaPE+WeakSup.\end{tabular}        & RetinaNet-Res50 & 33.53±0.17 & 53.23±0.09\\\midrule
\begin{tabular}[c]{@{}l@{}}Baseline:\\ Plain RGB-T Early Fusion\end{tabular}           & GFL-Res50       & 33.50±0.28 & 52.77±0.25 \\\midrule
\begin{tabular}[c]{@{}l@{}}Traditional Feature Distill+\\ Baseline+ShaPE+WeakSup.\end{tabular} & GFL-Res50       & 35.83±0.17 & 57.07±0.09 \\\midrule
\rowcolor[HTML]{D9D9D9} 
\begin{tabular}[c]{@{}l@{}}CoreKD+\\ Baseline+ShaPE+WeakSup.\end{tabular}        & GFL-Res50       & 37.03±0.09  & 57.70±0.08\\\bottomrule
\end{tabular}}}
\vspace{-1mm}
\end{table}

\begin{table}[t]
{
\caption{Comparison of traditional feature distillation with our CoreKD on the FLIR dataset~\cite{Flir}.}
\vspace{-1mm}
\label{table:comparisonKD_flir}
\setlength{\tabcolsep}{3pt}
\resizebox{\linewidth}{!}{
\begin{tabular}{l|l|l|l}
\toprule
Method                                                                & Detector        & mAP (↑)    & mAP50 (↑)  \\\midrule
\begin{tabular}[c]{@{}l@{}}Baseline:\\ Plain RGB-T Early Fusion\end{tabular}                   & RetinaNet-Res50 & 37.47±0.05 & 69.57±0.05 \\\midrule
\begin{tabular}[c]{@{}l@{}}Traditional Feature Distill+\\ Baseline+ShaPE+WeakSup.\end{tabular} & RetinaNet-Res50 & 38.10±0.14 & 70.97±0.12 \\\midrule
\rowcolor[HTML]{D9D9D9} 
\begin{tabular}[c]{@{}l@{}}CoreKD+\\ Baseline+ShaPE+WeakSup.\end{tabular}                      & RetinaNet-Res50 & 38.83±0.17 & 72.23±0.31 \\\midrule
\begin{tabular}[c]{@{}l@{}}Baseline:\\ Plain RGB-T Early Fusion\end{tabular}                   & GFL-Res50       & 41.90±0.22 & 74.77±0.17 \\\midrule
\begin{tabular}[c]{@{}l@{}}Traditional Feature Distill+\\ Baseline+ShaPE+WeakSup.\end{tabular} & GFL-Res50       & 43.63±0.09 & 77.80±0.08 \\\midrule
\rowcolor[HTML]{D9D9D9} 
\begin{tabular}[c]{@{}l@{}}CoreKD+\\ Baseline+ShaPE+WeakSup.\end{tabular}                      & GFL-Res50       & 44.00±0.00 & 78.17±0.05\\\bottomrule
\end{tabular}}}
\vspace{-3mm}
\end{table}

\textbf{Backbone Network.} We evaluate our EME method using ResNet-101~\cite{resnet} as the backbone network on the M3FD and FLIR datasets, and present the results in Table~\ref{table:comparisonBackbone}. We observe that detectors using ResNet-101 consistently achieve better performance than those using ResNet-50. For example, RetinaNet with ResNet-101 (74.87\%) obtains a 2.64\% mAP50 absolute gain over ResNet-50 (72.23\%).

\begin{table}[t]
{
\caption{Results of our EME method on the M3FD and FLIR datasets using different baseline detectors and backbone networks.}
\vspace{-1mm}
\label{table:comparisonBackbone}
\setlength{\tabcolsep}{5pt}
\renewcommand\arraystretch{1.2}
\resizebox{\linewidth}{!}{
\begin{tabular}{l|l|l|l|l|l}
\toprule
Datasets              & Detector  & Backbone & FLOPs   & mAP (↑)    & mAP50 (↑)  \\\midrule
\multirow{4}{*}{M3FD} & RetinaNet & ResNet-50        & 62.218G & 33.53±0.17 & 53.23±0.09 \\
& RetinaNet & ResNet-101       & 65.392G & 34.23±0.12 & 54.63±0.45 \\\cline{2-6}
& GFL       & ResNet-50        & 61.718G & 37.03±0.09 & 57.70±0.08 \\
& GFL       & ResNet-101       & 64.892G & 37.37±0.12 & 58.60±0.08 \\\midrule
\multirow{4}{*}{FLIR} & RetinaNet & ResNet-50        & 62.218G & 38.83±0.17 & 72.23±0.31 \\
& RetinaNet & ResNet-101       & 65.392G & 40.67±0.05 & 74.87±0.34 \\\cline{2-6}
& GFL       & ResNet-50        & 61.718G & 44.00±0.00 & 78.17±0.05 \\
& GFL       & ResNet-101       & 64.892G & 44.47±0.05 & 79.57±0.05\\\bottomrule
\end{tabular}}}
\vspace{-1mm}
\end{table}
}

%%%%%%%%%%%%%%%%%%% SOTA: Comparisons. M3FD + visualization + FLIR %%%%%%%%%%%%%%%%%%
\begin{figure*}
\begin{minipage}{\linewidth}
\begin{center}
\captionof{table}{Comparisons with state-of-the-art approaches on the M3FD dataset~\cite{liu2022target}. {The best results are highlighted in bold and marked in {\color{red}\textbf{red}}, while the second-best ones are underlined and marked in {\color[RGB]{0,128,0}\textbf{green}}. The detection results of our EME are obtained by running three independent experiments. The mean values and standard deviations of these results are reported.}}
\vspace{-1mm}
\label{table:sotam3fd}
\setlength{\tabcolsep}{3pt}
\resizebox{\linewidth}{!}{
\begin{tabular}{l|cccccccccccc>{\columncolor[HTML]{D9D9D9}}c}
\toprule
\multicolumn{14}{c}{(a) Dataset Splitting Method: Random Splitting}\\ \midrule
\multicolumn{1}{c|}{} & Thermal~\cite{yolov5} & RGB~\cite{yolov5}     & AUIF~\cite{AUIF}    & CDDF~\cite{CDDFuse} & DDcGAN~\cite{DDcGAN}  & DIVF~\cite{DIVFusion} & DenseF~\cite{DenseFuse} & PSF~\cite{PSFusion} & RFN~\cite{RFN-Nest}                              & SeAF~\cite{SeAFusion}                            & TarDAL~\cite{liu2022target}  & U2F~\cite{U2Fusion} & EME (Ours)                              \\ \midrule
mAP                   & 49.10 & 52.40 & 53.30 & 53.00 & 52.20 & 52.70   & 53.40   & 53.10  & {\color[HTML]{008000} \underline{53.50}} & 53.10                              & 52.50 & 53.40  & {\color[HTML]{FE0000} \textbf{54.00±0.28}} \\
mAP50                 & 77.30 & 81.90 & 81.90 & 80.90 & 81.60 & 81.50   & 81.70   & 82.00  & 81.70                              & {\color[HTML]{008000} \underline{82.20}} & 81.00 & 81.90  & {\color[HTML]{FE0000} \textbf{82.90±0.37}} \\ \midrule
Person                & 79.30 & 68.40 & 76.70 & 76.30 & 73.60 & 74.50   & 76.50   & 76.70  & 75.30                              & 77.00                              & 79.10 & 77.00  & {79.53±0.26}                                 \\
Car                   & 87.90 & 90.80 & 91.00 & 91.00 & 90.70 & 91.10   & 91.40   & 90.80  & 91.00                              & 91.10                              & 90.50 & 91.20  & {91.90±0.29}                                 \\
Bus                   & 87.20 & 92.20 & 90.00 & 90.10 & 90.70 & 91.60   & 89.40   & 90.10  & 89.40                              & 91.20                              & 89.40 & 90.70  & {89.80±0.45}                                 \\
Motor                 & 70.00 & 74.00 & 72.60 & 69.20 & 74.80 & 73.50   & 72.80   & 73.30  & 73.30                              & 72.20                              & 70.30 & 71.30  & {74.87±0.95}                                 \\
TrafficLight          & 55.90 & 80.30 & 77.40 & 75.40 & 76.90 & 74.80   & 77.20   & 78.20  & 77.40                              & 77.60                              & 72.70 & 77.70  & {77.40±1.13}                                 \\
Truck                 & 83.40 & 85.70 & 83.70 & 83.10 & 82.90 & 83.40   & 82.90   & 82.90  & 83.90                              & 84.10                              & 84.00 & 83.60  & {84.00±0.93}                                 \\
\midrule
\multicolumn{14}{c}{(b) Dataset Splitting Method: M3FD-zxSplit}\\ \midrule
\multicolumn{1}{c|}{}& Thermal~\cite{yolov5} & RGB~\cite{yolov5}     & AUIF~\cite{AUIF}    & CDDF~\cite{CDDFuse} & DDcGAN~\cite{DDcGAN}  & DIVF~\cite{DIVFusion} & DenseF~\cite{DenseFuse} & PSF~\cite{PSFusion} & RFN~\cite{RFN-Nest}                              & SeAF~\cite{SeAFusion}                            & TarDAL~\cite{liu2022target}  & U2F~\cite{U2Fusion} & EME (Ours)                               \\ \midrule
mAP                   & 34.90 & 36.10 & 38.30 & 38.60 & 37.10 & 37.10 & 38.90                              & 38.00 & {\color[HTML]{000000} 38.20} & 38.90                        & {\color[HTML]{008000} \underline{39.10}} & 38.70 & {\color[HTML]{FE0000} \textbf{41.10±0.29}} \\
mAP50                 & 57.20 & 60.20 & 62.00 & 61.90 & 61.00 & 60.80 & {\color[HTML]{008000} \underline{62.40}} & 61.10 & 61.30                        & {\color[HTML]{000000} 62.20} & 61.90                              & 61.90 & {\color[HTML]{FE0000} \textbf{66.23±0.40}} \\ \midrule
Person                & 74.60 & 55.90 & 72.20 & 71.90 & 67.30 & 67.60 & 72.30                              & 71.70 & 70.50                        & 72.50                        & 75.50                              & 72.40 & {77.23±0.26}                                 \\
Car                   & 80.20 & 84.80 & 85.50 & 85.60 & 84.90 & 85.20 & 85.90                              & 85.50 & 85.80                        & 85.50                        & 85.00                              & 85.50 & {87.13±0.19}                                 \\
Bus                   & 58.30 & 65.70 & 58.60 & 61.80 & 61.60 & 59.80 & 61.40                              & 58.30 & 61.30                        & 61.50                        & 60.90                              & 60.10 & {62.33±1.96}                                 \\
Motor                 & 48.00 & 45.10 & 49.10 & 47.60 & 49.00 & 48.70 & 49.60                              & 45.80 & 44.60                        & 47.50                        & 46.80                              & 50.80 & {55.33±0.21}                                 \\
TrafficLight          & 27.30 & 56.80 & 49.80 & 48.70 & 49.10 & 51.20 & 48.60                              & 50.90 & 49.70                        & 50.80                        & 46.90                              & 48.00 & {55.33±0.37}                                 \\
Truck                 & 54.80 & 52.70 & 56.70 & 55.50 & 53.80 & 52.60 & 56.80                              & 54.70 & 55.80                        & 55.70                        & 56.70                              & 54.70 & {60.10±0.16}                                 \\ \bottomrule
\end{tabular}}
\end{center}
\end{minipage}
\begin{minipage}{\linewidth}
\vspace{1mm}
\includegraphics[width=\linewidth]{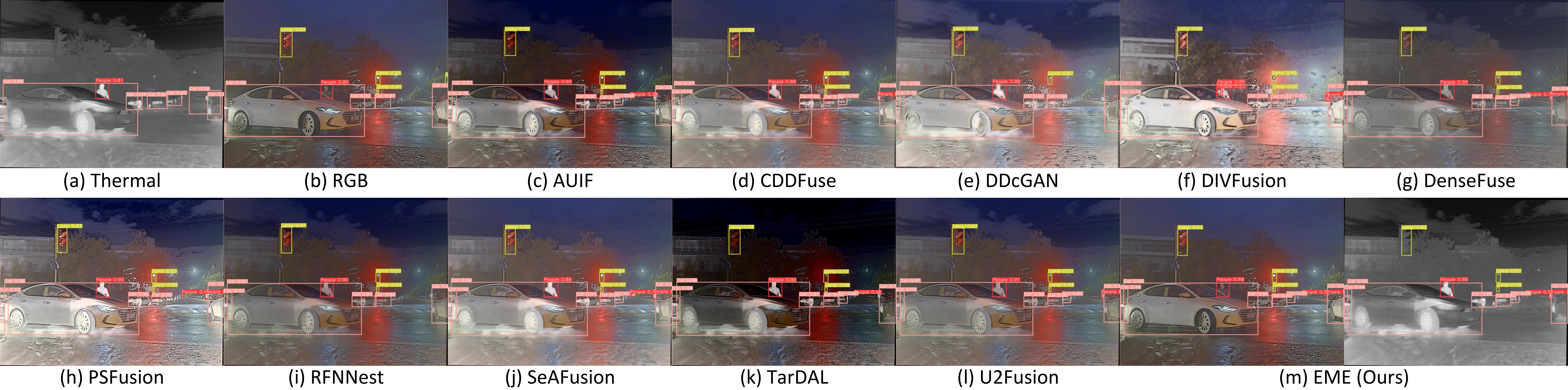}
\caption{Detection results of the YOLOv5~\cite{yolov5} detector on one example scene from the M3FD~\cite{liu2022target} dataset. (a) and (b) respectively show the results using only a thermal image and only an RGB image. (c)-(l) display the detection results using fused images obtained from 10 different image fusion approaches. (m) demonstrates the results using our EME method.}
\vspace{-2mm}
\label{fig:comparisonsotam3fd}
\end{minipage}
\begin{minipage}{\linewidth}
\vspace{5mm}
\begin{center}
\captionof{table}{Comparisons with state-of-the-art approaches on the FLIR~\cite{Flir} dataset. {The best results are highlighted in bold and marked in {\color{red}\textbf{red}}, while the second-best ones are underlined and marked in {\color[RGB]{0,128,0}\textbf{green}}. The detection results of our EME are obtained by running three independent experiments. The mean values and standard deviations of these results are reported.}}
\vspace{-1mm}
\label{table:sotaflir}
\setlength{\tabcolsep}{3pt}
\resizebox{\linewidth}{!}{
\begin{tabular}{l|ccccccccccccc>{\columncolor[HTML]{D9D9D9}}c }
\toprule
& CBF~\cite{Woo_2018_ECCV} & MCG~\cite{devaguptapu2019borrow} & MUN~\cite{devaguptapu2019borrow} & ODS~\cite{Munir2020ThermalOD}  & CFR~\cite{CFR}   & GAFF~\cite{zhang2021guided}  & BU~\cite{BU} & SMPD~\cite{SMPD}  & ThDe~\cite{thermaldet} & MSAT~\cite{MSAT}  & CSAA~\cite{CSAA}  & MFPT~\cite{MFPT}  & ProbEn3~\cite{proben} & EME (Ours)    \\ \midrule
mAP50   & 67.20      & 61.40    & 61.54      & 69.62 & 72.39 & 72.90 & 73.20    & 73.58 & 74.60      & 76.20 & 79.20 & 80.00 & {\color[RGB]{0,128,0}\underline{83.76}}   & \textbf{\color{red}84.63±0.12} \\ \midrule
Bicycle & 60.50      & 50.26    & 49.43      & 55.53 & 55.77 & -     & 57.40    & 56.20 & 60.04      & -     & -     & 67.70 & 73.49   &{79.73±0.09}               \\
Car     & 83.60      & 70.63    & 70.72      & 82.33 & 84.91 & -     & 86.50    & 85.80 & 85.52      & -     & -     & 89.00 & 90.14   & {92.67±0.12}              \\
Person  & 57.60      & 63.31    & 64.47      & 71.01 & 74.49 & -     & 75.60    & 78.74 & 78.24      & -     & -     & 83.20 & 87.65   & {81.27±0.09}              \\ \bottomrule
\end{tabular}}
\end{center}
\end{minipage}
\vspace{-4mm}
\end{figure*}
%%%%%%%%%%%%%%%%%%%%%%%%%%%%%%%%%%%%%%%%%%%%%%%%%%%%%%%%%%%%%

\subsection{Comparison with the State-of-The-Art Approaches}\label{sec:sota}
We use the one-stage YOLOv5~\cite{yolov5} detector as the baseline, and incorporate our proposed modules to construct the effective multispectral early-fusion (EME) model. Table~\ref{table:sotam3fd} and Table~\ref{table:sotaflir} compares our EME and previous state-of-the-art approaches on M3FD~\cite{liu2022target} and FLIR~\cite{Flir} datasets.

In Table~\ref{table:sotam3fd}, we compare our EME with 10 state-of-the-art image-fusion-based object detection approaches~\cite{AUIF, CDDFuse, DDcGAN, DIVFusion, DenseFuse, PSFusion, RFN-Nest, SeAFusion, liu2022target, U2Fusion}. We first generate fused images based on their official implementations, and then train YOLOv5~\cite{yolov5} using these fused images with the same training settings. The results show that our EME achieves state-of-the-art performance. We observe that the results in Table~\ref{table:sotam3fd}~(a) are obviously better than those in Table~\ref{table:sotam3fd}~(b). This demonstrates that random splitting causes information leakage and makes it difficult to improve performance. Fig.~\ref{fig:comparisonsotam3fd} presents an example scene for visualization. {Compared to other approaches, a weakness of our EME detector is that it doesn't generate a fused image for direct visualization. This is because our method focuses on detection rather than image fusion. We will address this issue in future work.}

In Table~\ref{table:sotaflir}, we compare our EME with 13 multispectral object detection approaches. These approaches include (1) medium-fusion strategies, such as CBF~\cite{Woo_2018_ECCV}, MCG~\cite{devaguptapu2019borrow}, MUN~\cite{devaguptapu2019borrow}, CFR~\cite{CFR}, GAFF~\cite{zhang2021guided}, SMPD~\cite{SMPD}, MSAT~\cite{MSAT}, CSAA~\cite{CSAA}, and MFPT~\cite{MFPT}; (2) domain adaptation and single-modality detection approaches, such as ODS~\cite{Munir2020ThermalOD}, BU~\cite{BU}, and ThDe~\cite{thermaldet}; and (3) late-fusion strategy~\cite{proben}. The results show that our EME also achieves state-of-the-art performance on the FLIR dataset~\cite{Flir}.

{
\subsection{Comparison of Inference Efficiency}
We compare the inference efficiency of our EME method with previous state-of-the-art approaches on an edge device: the NVIDIA AGX Orin with 64GB of memory. We select open-source approaches for comparison and adopt YOLOv5-small as the baseline detector for all methods.

Table~\ref{table:efficiencyComparison} presents the FLOPs, number of parameters, and inference time for various methods. Experimental results show that our EME method is the fastest. Interestingly, we notice that a reduction in FLOPs does not directly lead to a similar reduction in the inference time of an approach. This phenomenon may be attributed to the frequent memory access by operators, as confirmed in PConv\cite{pconv}. This observation inspires us to further speed up our EME method by reducing memory access in the future.

\begin{table}[t]
{
\caption{Comparison of inference efficiency. All methods use the YOLOv5-small as the baseline detector. Inference time is evaluated on an edge device: the NVIDIA AGX Orin.}
\vspace{-1mm}
\label{table:efficiencyComparison}
\setlength{\tabcolsep}{10pt}
\resizebox{\linewidth}{!}{
\begin{tabular}{l|l|l|l}
\toprule
{Method}     & {FLOPs}     & {Parameters} & {Time (seconds)} \\
\midrule
{AUIF\cite{AUIF}}       & {12.185G}   & {7.037M}     & {5.12s}          \\
\midrule
{CDDF\cite{CDDFuse}}       & {2816.279G} & {8.214M}     & {9.507s}         \\
\midrule
{DensF\cite{DenseFuse}}      & {151.596G}  & {7.100M}     & {1.769s}         \\
\midrule
{PSF\cite{PSFusion}}        & {939.024G}  & {52.925M}    & {2.512s}         \\
\midrule
{RFN\cite{RFN-Nest}}        & {1859.908G} & {9.759M}     & {3.473s}         \\
\midrule
{SeAF\cite{SeAFusion}}       & {272.945G}  & {7.192M}     & {1.344s}         \\
\midrule
{TarDAL\cite{liu2022target}}     & {478.474G}  & {7.323M}     & {1.446s}         \\
\midrule
\rowcolor[HTML]{D9D9D9} 
{EME (Ours)} & {15.780G}   & {7.063M}     & {0.077s}\\\bottomrule
\end{tabular}}}
\vspace{-5mm}
\end{table}
}

\section{Conclusions}
In this paper, we propose the effective multispectral early-fusion (EME) detector, which achieves both high performance and efficiency. We identify and address performance obstacles in a plain early-fusion strategy, such as information interference, domain gaps, and weak feature representation, by proposing solutions including shape-priority early-fusion modules, weakly supervised learning, and core knowledge distillation. Extensive experiments on representative datasets demonstrate the effectiveness and efficiency of our EME detector.

{The main advantage of our EME detector is that it improves the performance of an efficient single-branch early-fusion strategy without significantly increasing its computational burden. We demonstrate that our EME detector has similar FLOPs and parameters to a plain early-fusion strategy, while achieving better performance than a cumbersome two-branch structure. We also show that our EME detector has higher inference efficiency than the two-branch structure on an edge device: the NVIDIA AGX Orin.

A limitation of our current EME detector is its inefficient two-stage training paradigm in the knowledge distillation technique. In the future, we will work towards an optimized one-stage paradigm to accelerate the training process and further improve detection accuracy.
}

\bibliographystyle{IEEEtran}
\bibliography{zxbib.bib}

\begin{thebibliography}{10}
\providecommand{\url}[1]{#1}
\csname url@samestyle\endcsname
\providecommand{\newblock}{\relax}
\providecommand{\bibinfo}[2]{#2}
\providecommand{\BIBentrySTDinterwordspacing}{\spaceskip=0pt\relax}
\providecommand{\BIBentryALTinterwordstretchfactor}{4}
\providecommand{\BIBentryALTinterwordspacing}{\spaceskip=\fontdimen2\font plus
\BIBentryALTinterwordstretchfactor\fontdimen3\font minus
  \fontdimen4\font\relax}
\providecommand{\BIBforeignlanguage}[2]{{%
\expandafter\ifx\csname l@#1\endcsname\relax
\typeout{** WARNING: IEEEtran.bst: No hyphenation pattern has been}%
\typeout{** loaded for the language `#1'. Using the pattern for}%
\typeout{** the default language instead.}%
\else
\language=\csname l@#1\endcsname
\fi
#2}}
\providecommand{\BIBdecl}{\relax}
\BIBdecl

\bibitem{liu2022target}
J.~Liu, X.~Fan, Z.~Huang, G.~Wu, R.~Liu, W.~Zhong, and Z.~Luo, ``{Target-Aware
  Dual Adversarial Learning and a Multi-Scenario Multi-Modality Benchmark to
  Fuse Infrared and Visible for Object Detection},'' in \emph{Proceedings of
  the Conference on Computer Vision and Pattern Recognition}, 2022.

\bibitem{yolov5}
\BIBentryALTinterwordspacing
G.~Jocher, ``{YOLOv5 by Ultralytics},'' 2020. [Online]. Available:
  \url{https://github.com/ultralytics/yolov5}
\BIBentrySTDinterwordspacing

\bibitem{pdmsc}
Z.~Chen and X.~Huang, ``{Pedestrian Detection for Autonomous Vehicle Using
  Multi-Spectral Cameras},'' \emph{IEEE Transactions on Intelligent Vehicles},
  vol.~4, no.~2, pp. 211--219, 2019.

\bibitem{CACFNet}
W.~Zhou, S.~Dong, M.~Fang, and L.~Yu, ``{CACFNet: Cross-Modal Attention
  Cascaded Fusion Network for RGB-T Urban Scene Parsing},'' \emph{IEEE
  Transactions on Intelligent Vehicles}, vol.~9, no.~1, pp. 1919--1929, 2024.

\bibitem{RBIT}
Y.~Liu, C.~Hu, B.~Zhao, Y.~Huang, and X.~Zhang, ``{Region-Based
  Illumination-Temperature Awareness and Cross-Modality Enhancement for
  Multispectral Pedestrian Detection},'' \emph{IEEE Transactions on Intelligent
  Vehicles}, pp. 1--12, 2024.

\bibitem{farooq2022evaluation}
M.~A. Farooq, W.~Shariff, and P.~Corcoran, ``{Evaluation of Thermal Imaging on
  Embedded GPU Platforms for Application in Vehicular Assistance Systems},''
  \emph{IEEE Transactions on Intelligent Vehicles}, vol.~8, no.~2, pp.
  1130--1144, 2022.

\bibitem{TISPT}
M.~Ding, W.-H. Chen, and Y.-F. Cao, ``{Thermal Infrared Single-Pedestrian
  Tracking for Advanced Driver Assistance System},'' \emph{IEEE Transactions on
  Intelligent Vehicles}, vol.~8, no.~1, pp. 814--824, 2023.

\bibitem{MTANet}
W.~Zhou, S.~Dong, J.~Lei, and L.~Yu, ``{MTANet: Multitask-Aware Network With
  Hierarchical Multimodal Fusion for RGB-T Urban Scene Understanding},''
  \emph{IEEE Transactions on Intelligent Vehicles}, vol.~8, no.~1, pp. 48--58,
  2023.

\bibitem{umssde}
Y.~Guo, H.~Kong, and S.~Gu, ``{Unsupervised Multi-Spectrum Stereo Depth
  Estimation for All-Day Vision},'' \emph{IEEE Transactions on Intelligent
  Vehicles}, vol.~9, no.~1, pp. 501--511, 2024.

\bibitem{qafa}
Y.~Zhu, C.~Li, J.~Tang, and B.~Luo, ``{Quality-Aware Feature Aggregation
  Network for Robust RGB-T Tracking},'' \emph{IEEE Transactions on Intelligent
  Vehicles}, vol.~6, no.~1, pp. 121--130, 2021.

\bibitem{liu2016multispectral}
J.~Liu, S.~Zhang, S.~Wang, and D.~N. Metaxas, ``{Multispectral Deep Neural
  Networks for Pedestrian Detection},'' in \emph{Proceedings of the British
  Machine Vision Conference}, 2016.

\bibitem{wagner2016multispectral}
J.~Wagner, V.~Fischer, M.~Herman, S.~Behnke \emph{et~al.}, ``{Multispectral
  Pedestrian Detection using Deep Fusion Convolutional Neural Networks},'' in
  \emph{Proceedings of the European Symposium on Artificial Neural Networks},
  vol. 587, 2016, pp. 509--514.

\bibitem{xie2023illumination}
Q.~Xie, T.-Y. Cheng, Z.~Dai, V.~Tran, N.~Trigoni, and A.~Markham,
  ``{Illumination-Aware Hallucination-Based Domain Adaptation for Thermal
  Pedestrian Detection},'' \emph{IEEE Transactions on Intelligent
  Transportation Systems}, 2023.

\bibitem{liu2021deep}
T.~Liu, K.-M. Lam, R.~Zhao, and G.~Qiu, ``{Deep Cross-Modal Representation
  Learning and Distillation for Illumination-Invariant Pedestrian Detection},''
  \emph{IEEE Transactions on Circuits and Systems for Video Technology},
  vol.~32, no.~1, pp. 315--329, 2021.

\bibitem{zhang2021guided}
H.~Zhang, E.~Fromont, S.~Lef{\`e}vre, and B.~Avignon, ``{Guided Attentive
  Feature Fusion for Multispectral Pedestrian Detection},'' in
  \emph{Proceedings of the Winter Conference on Applications of Computer
  Vision}, 2021, pp. 72--80.

\bibitem{dformer}
B.~Yin, X.~Zhang, Z.~Li, L.~Liu, M.-M. Cheng, and Q.~Hou, ``{DFormer:
  Rethinking RGBD Representation Learning for Semantic Segmentation},'' in
  \emph{Proceedings of the International Conference on Learning
  Representations}, 2024.

\bibitem{clip}
A.~Radford, J.~W. Kim, C.~Hallacy, A.~Ramesh, G.~Goh, S.~Agarwal, G.~Sastry,
  A.~Askell, P.~Mishkin, J.~Clark, G.~Krueger, and I.~Sutskever, ``{Learning
  Transferable Visual Models from Natural Language Supervision},'' in
  \emph{Proceedings of the International Conference on Machine Learning}, vol.
  139, 2021, pp. 8748--8763.

\bibitem{PADCLIP}
Z.~Lai, N.~Vesdapunt, N.~Zhou, J.~Wu, C.~P. Huynh, X.~Li, K.~K. Fu, and C.-N.
  Chuah, ``{PADCLIP: Pseudo-Labeling with Adaptive Debiasing in CLIP for
  Unsupervised Domain Adaptation},'' in \emph{Proceedings of the International
  Conference on Computer Vision}, 2023, pp. 16\,109--16\,119.

\bibitem{hinton2015distilling}
G.~Hinton, O.~Vinyals, and J.~Dean, ``{Distilling the Knowledge in a Neural
  Network},'' in \emph{Proceedings of the Advances in Neural Information
  Processing Systems Workshop}, 2015.

\bibitem{fitnet}
A.~Romero, N.~Ballas, S.~E. Kahou, A.~Chassang, C.~Gatta, and Y.~Bengio,
  ``{FitNets: Hints for Thin Deep Nets},'' in \emph{Proceedings of the
  International Conference on Learning Representations}, 2015.

\bibitem{chen2017learning}
G.~Chen, W.~Choi, X.~Yu, T.~Han, and M.~Chandraker, ``{Learning Efficient
  Object Detection Models with Knowledge Distillation},'' \emph{Advances in
  Neural Information Processing Systems}, vol.~30, 2017.

\bibitem{lin2017focal}
T.-Y. Lin, P.~Goyal, R.~Girshick, K.~He, and P.~Doll{\'a}r, ``{Focal Loss for
  Dense Object Detection},'' in \emph{Proceedings of the International
  Conference on Computer Vision}, 2017, pp. 2980--2988.

\bibitem{gfl}
X.~Li, W.~Wang, L.~Wu, S.~Chen, X.~Hu, J.~Li, J.~Tang, and J.~Yang,
  ``{Generalized Focal Loss: Learning Qualified and Distributed Bounding Boxes
  for Dense Object Detection},'' \emph{Advances in Neural Information
  Processing Systems}, vol.~33, pp. 21\,002--21\,012, 2020.

\bibitem{Flexible-Mixup}
J.~Wang, M.~Zhang, W.~Li, and R.~Tao, ``{A Multistage Information Complementary
  Fusion Network Based on Flexible-Mixup for HSI-X Image Classification},''
  \emph{IEEE Transactions on Neural Networks and Learning Systems}, pp. 1--13,
  2023.

\bibitem{mifn}
J.~Wang, W.~Li, Y.~Gao, M.~Zhang, R.~Tao, and Q.~Du, ``{Hyperspectral and SAR
  Image Classification via Multiscale Interactive Fusion Network},'' \emph{IEEE
  Transactions on Neural Networks and Learning Systems}, vol.~34, no.~12, pp.
  10\,823--10\,837, 2023.

\bibitem{slanet}
M.~Zhang, W.~Li, X.~Zhao, H.~Liu, R.~Tao, and Q.~Du, ``{Morphological
  Transformation and Spatial-Logical Aggregation for Tree Species
  Classification Using Hyperspectral Imagery},'' \emph{IEEE Transactions on
  Geoscience and Remote Sensing}, vol.~61, pp. 1--12, 2023.

\bibitem{csma}
Y.~Gao, M.~Zhang, J.~Wang, and W.~Li, ``{Cross-Scale Mixing Attention for
  Multisource Remote Sensing Data Fusion and Classification},'' \emph{IEEE
  Transactions on Geoscience and Remote Sensing}, vol.~61, pp. 1--15, 2023.

\bibitem{proben}
Y.-T. Chen, J.~Shi, Z.~Ye, C.~Mertz, D.~Ramanan, and S.~Kong, ``{Multimodal
  Object Detection via Probabilistic Ensembling},'' in \emph{Proceedings of the
  European Conference on Computer Vision}, 2022, pp. 139--158.

\bibitem{zhang2022low}
H.~Zhang, E.~Fromont, S.~Lef{\`e}vre, and B.~Avignon, ``{Low-Cost Multispectral
  Scene Analysis with Modality Distillation},'' in \emph{Proceedings of the
  Winter Conference on Applications of Computer Vision}, 2022, pp. 803--812.

\bibitem{CDDFuse}
Z.~Zhao, H.~Bai, J.~Zhang, Y.~Zhang, S.~Xu, Z.~Lin, R.~Timofte, and
  L.~Van~Gool, ``{CDDFuse: Correlation-Driven Dual-Branch Feature Decomposition
  for Multi-Modality Image Fusion},'' in \emph{Proceedings of the Conference on
  Computer Vision and Pattern Recognition}, 2023, pp. 5906--5916.

\bibitem{DIVFusion}
L.~Tang, X.~Xiang, H.~Zhang, M.~Gong, and J.~Ma, ``{DIVFusion: Darkness-Free
  Infrared and Visible Image Fusion},'' \emph{Information Fusion}, vol.~91, pp.
  477--493, 2023.

\bibitem{zhang2021weakly}
D.~Zhang, J.~Han, G.~Cheng, and M.-H. Yang, ``{Weakly Supervised Object
  Localization and Detection: A Survey},'' \emph{IEEE Transactions on Pattern
  Analysis and Machine Intelligence}, vol.~44, no.~9, pp. 5866--5885, 2021.

\bibitem{zhang2023illumination}
Y.~Zhang, H.~Yu, Y.~He, X.~Wang, and W.~Yang, ``{Illumination-Guided RGBT
  Object Detection with Inter- and Intra-Modality Fusion},'' \emph{IEEE
  Transactions on Instrumentation and Measurement}, vol.~72, pp. 1--13, 2023.

\bibitem{zhang2023tfdet}
X.~Zhang, X.~Zhang, Z.~Sheng, and H.-L. Shen, ``{TFDet: Target-Aware Fusion for
  RGB-T Pedestrian Detection},'' \emph{arXiv preprint arXiv:2305.16580}, 2023.

\bibitem{li2023object}
Z.~Li, P.~Xu, X.~Chang, L.~Yang, Y.~Zhang, L.~Yao, and X.~Chen, ``{When Object
  Detection Meets Knowledge Distillation: A Survey},'' \emph{IEEE Transactions
  on Pattern Analysis and Machine Intelligence}, 2023.

\bibitem{SSIM}
Z.~Wang, A.~C. Bovik, H.~R. Sheikh, and E.~P. Simoncelli, ``{Image Quality
  Assessment: From Error Visibility to Structural Similarity},'' \emph{IEEE
  Transactions on Image Processing}, vol.~13, no.~4, pp. 600--612, 2004.

\bibitem{ILSVRC15}
O.~Russakovsky, J.~Deng, H.~Su, J.~Krause, S.~Satheesh, S.~Ma, Z.~Huang,
  A.~Karpathy, A.~Khosla, M.~Bernstein, A.~C. Berg, and L.~Fei-Fei, ``{ImageNet
  Large Scale Visual Recognition Challenge},'' \emph{International Journal of
  Computer Vision}, vol. 115, no.~3, pp. 211--252, 2015.

\bibitem{lin2014microsoft}
T.-Y. Lin, M.~Maire, S.~Belongie, J.~Hays, P.~Perona, D.~Ramanan,
  P.~Doll{\'a}r, and C.~L. Zitnick, ``{Microsoft COCO: Common Objects in
  Context},'' in \emph{Proceedings of the European Conference on Computer
  Vision}, 2014, pp. 740--755.

\bibitem{Flir}
``{FREE FLIR Thermal Dataset for Algorithm Training},''
  \url{https://www.flir.com/oem/adas/adas-dataset-form/}.

\bibitem{unveilingclip}
Z.~Chen, Z.~Zhang, X.~Tan, Y.~Qu, and Y.~Xie, ``{Unveiling the Power of CLIP in
  Unsupervised Visible-Infrared Person Re-Identification},'' in
  \emph{Proceedings of the ACM International Conference on Multimedia}, 2023,
  pp. 3667--3675.

\bibitem{textif}
X.~Yi, H.~Xu, H.~Zhang, L.~Tang, and J.~Ma, ``{Text-IF: Leveraging Semantic
  Text Guidance for Degradation-Aware and Interactive Image Fusion},'' in
  \emph{Proceedings of the Conference on Computer Vision and Pattern
  Recognition}, 2024, pp. 27\,026--27\,035.

\bibitem{clipdrivenperson}
\BIBentryALTinterwordspacing
X.~Yu, N.~Dong, L.~Zhu, H.~Peng, and D.~Tao, ``{CLIP-Driven Semantic Discovery
  Network for Visible-Infrared Person Re-Identification},'' 2024. [Online].
  Available: \url{https://arxiv.org/abs/2401.05806}
\BIBentrySTDinterwordspacing

\bibitem{unidetector}
Z.~Wang, Y.~Li, X.~Chen, S.-N. Lim, A.~Torralba, H.~Zhao, and S.~Wang,
  ``{Detecting Everything in the Open World: Towards Universal Object
  Detection},'' in \emph{Proceedings of the Conference on Computer Vision and
  Pattern Recognition}, June 2023, pp. 11\,433--11\,443.

\bibitem{cdul}
R.~Abdelfattah, Q.~Guo, X.~Li, X.~Wang, and S.~Wang, ``{CDUL: CLIP-Driven
  Unsupervised Learning for Multi-Label Image Classification},'' in
  \emph{Proceedings of the International Conference on Computer Vision}, 2023,
  pp. 1348--1357.

\bibitem{regionclip}
Y.~Zhong, J.~Yang, P.~Zhang, C.~Li, N.~Codella, L.~H. Li, L.~Zhou, X.~Dai,
  L.~Yuan, Y.~Li, and J.~Gao, ``{RegionCLIP: Region-Based Language-Image
  Pretraining},'' in \emph{Proceedings of the Conference on Computer Vision and
  Pattern Recognition}, June 2022, pp. 16\,793--16\,803.

\bibitem{mutual}
Y.~Zhang, T.~Xiang, T.~M. Hospedales, and H.~Lu, ``{Deep Mutual Learning},'' in
  \emph{Proceedings of the Conference on Computer Vision and Pattern
  Recognition}, 2018, pp. 4320--4328.

\bibitem{kdlsr}
L.~Yuan, F.~E. Tay, G.~Li, T.~Wang, and J.~Feng, ``{Revisiting Knowledge
  Distillation via Label Smoothing Regularization},'' in \emph{Proceedings of
  the Conference on Computer Vision and Pattern Recognition}, 2020, pp.
  3903--3911.

\bibitem{resnet}
K.~He, X.~Zhang, S.~Ren, and J.~Sun, ``{Deep Residual Learning for Image
  Recognition},'' in \emph{Proceedings of the Conference on Computer Vision and
  Pattern Recognition}, 2016, pp. 770--778.

\bibitem{alignedFLIR}
V.~Sam, K.~Ali, M.~Christian, K.~Laurent, and E.~Lutz, ``{Robust Environment
  Perception for Automated Driving: A Unified Learning Pipeline for
  Visual-Infrared Object Detection},'' in \emph{IEEE Intelligent Vehicles
  Symposium}, 2022, pp. 367--374.

\bibitem{mmdetection}
\BIBentryALTinterwordspacing
{MMDetection Contributors}, ``{OpenMMLab Detection Toolbox and Benchmark},''
  2018. [Online]. Available: \url{https://github.com/open-mmlab/mmdetection}
\BIBentrySTDinterwordspacing

\bibitem{AUIF}
Z.~Zhao, S.~Xu, J.~Zhang, C.~Liang, C.~Zhang, and J.~Liu, ``{Efficient and
  Model-Based Infrared and Visible Image Fusion via Algorithm Unrolling},''
  \emph{IEEE Transactions on Circuits and Systems for Video Technology},
  vol.~32, no.~3, pp. 1186--1196, 2022.

\bibitem{DDcGAN}
J.~Ma, H.~Xu, J.~Jiang, X.~Mei, and X.-P. Zhang, ``{DDcGAN: A
  Dual-Discriminator Conditional Generative Adversarial Network for
  Multi-Resolution Image Fusion},'' \emph{IEEE Transactions on Image
  Processing}, vol.~29, pp. 4980--4995, 2020.

\bibitem{DenseFuse}
H.~Li and X.-J. Wu, ``{DenseFuse: A Fusion Approach to Infrared and Visible
  Images},'' \emph{IEEE Transactions on Image Processing}, vol.~28, no.~5, pp.
  2614--2623, 2019.

\bibitem{PSFusion}
L.~Tang, H.~Zhang, H.~Xu, and J.~Ma, ``{Rethinking the Necessity of Image
  Fusion in High-Level Vision Tasks: A Practical Infrared and Visible Image
  Fusion Network Based on Progressive Semantic Injection and Scene Fidelity},''
  \emph{Information Fusion}, vol.~99, p. 101870, 2023.

\bibitem{RFN-Nest}
H.~Li, X.-J. Wu, and J.~Kittler, ``{RFN-Nest: An End-to-End Residual Fusion
  Network for Infrared and Visible Images},'' \emph{Information Fusion},
  vol.~73, pp. 72--86, 2021.

\bibitem{SeAFusion}
L.~Tang, J.~Yuan, and J.~Ma, ``{Image Fusion in the Loop of High-Level Vision
  Tasks: A Semantic-Aware Real-Time Infrared and Visible Image Fusion
  Network},'' \emph{Information Fusion}, vol.~82, pp. 28--42, 2022.

\bibitem{U2Fusion}
H.~Xu, J.~Ma, J.~Jiang, X.~Guo, and H.~Ling, ``{U2Fusion: A Unified
  Unsupervised Image Fusion Network},'' \emph{IEEE Transactions on Pattern
  Analysis and Machine Intelligence}, 2020.

\bibitem{Woo_2018_ECCV}
S.~Woo, J.~Park, J.-Y. Lee, and I.~S. Kweon, ``{CBAM: Convolutional Block
  Attention Module},'' in \emph{Proceedings of the European Conference on
  Computer Vision}, 2018.

\bibitem{devaguptapu2019borrow}
C.~Devaguptapu, N.~Akolekar, M.~M~Sharma, and V.~N~Balasubramanian, ``{Borrow
  from Anywhere: Pseudo Multi-Modal Object Detection in Thermal Imagery},'' in
  \emph{Proceedings of the Conference on Computer Vision and Pattern
  Recognition Workshops}, 2019, pp. 0--0.

\bibitem{Munir2020ThermalOD}
\BIBentryALTinterwordspacing
F.~Munir, S.~Azam, M.~A. Rafique, A.~M. Sheri, and M.~Jeon, ``{Thermal Object
  Detection using Domain Adaptation through Style Consistency},'' \emph{ArXiv},
  vol. abs/2006.00821, 2020. [Online]. Available:
  \url{https://api.semanticscholar.org/CorpusID:219176719}
\BIBentrySTDinterwordspacing

\bibitem{CFR}
H.~Zhang, E.~Fromont, S.~Lefevre, and B.~Avignon, ``{Multispectral Fusion for
  Object Detection with Cyclic Fuse-and-Refine Blocks},'' in \emph{Proceedings
  of the International Conference on Image Processing}, 2020, pp. 276--280.

\bibitem{BU}
M.~Kieu, A.~D. Bagdanov, and M.~Bertini, ``{Bottom-Up and Layerwise Domain
  Adaptation for Pedestrian Detection in Thermal Images},'' \emph{ACM
  Transactions on Multimedia Computing, Communications, and Applications},
  vol.~17, no.~1, 2021.

\bibitem{SMPD}
Q.~Li, C.~Zhang, Q.~Hu, P.~Zhu, H.~Fu, and L.~Chen, ``{Stabilizing
  Multispectral Pedestrian Detection with Evidential Hybrid Fusion},''
  \emph{IEEE Transactions on Circuits and Systems for Video Technology},
  vol.~34, no.~4, pp. 3017--3029, 2024.

\bibitem{thermaldet}
Y.~Cao, T.~Zhou, X.~Zhu, and Y.~Su, ``{Every Feature Counts: An Improved
  One-Stage Detector in Thermal Imagery},'' in \emph{Proceedings of the
  International Conference on Computer and Communications}, 2019, pp.
  1965--1969.

\bibitem{MSAT}
S.~You, X.~Xie, Y.~Feng, C.~Mei, and Y.~Ji, ``{Multi-Scale Aggregation
  Transformers for Multispectral Object Detection},'' \emph{IEEE Signal
  Processing Letters}, vol.~30, pp. 1172--1176, 2023.

\bibitem{CSAA}
Y.~Cao, J.~Bin, J.~Hamari, E.~Blasch, and Z.~Liu, ``{Multimodal Object
  Detection by Channel Switching and Spatial Attention},'' in \emph{Proceedings
  of the Conference on Computer Vision and Pattern Recognition Workshops},
  2023, pp. 403--411.

\bibitem{MFPT}
Y.~Zhu, X.~Sun, M.~Wang, and H.~Huang, ``{Multi-Modal Feature Pyramid
  Transformer for RGB-Infrared Object Detection},'' \emph{IEEE Transactions on
  Intelligent Transportation Systems}, vol.~24, no.~9, pp. 9984--9995, 2023.

\bibitem{pconv}
J.~Chen, S.-h. Kao, H.~He, W.~Zhuo, S.~Wen, C.-H. Lee, and S.-H.~G. Chan,
  ``{Run, Don't Walk: Chasing Higher FLOPS for Faster Neural Networks},'' in
  \emph{Proceedings of the Conference on Computer Vision and Pattern
  Recognition}, June 2023, pp. 12\,021--12\,031.

\end{thebibliography}

\end{document}